%% file: main.tex
\documentclass[10pt,twocolumn,letterpaper,dvipsnames,svgnames]{article}

\usepackage{iccv}                


\usepackage{amsmath}
\usepackage{amssymb}
\usepackage{mathtools}
\usepackage{algorithm}
\usepackage{algorithmic}

\usepackage{times}
\usepackage{fix-cm}
\usepackage{soul}
\usepackage{url}
\usepackage{ragged2e}

\usepackage{graphicx}
\usepackage{epsfig}
\usepackage{subcaption}

\usepackage{array}
\usepackage{booktabs}
\usepackage{multirow}
\usepackage{makecell}

\usepackage{color}      
\usepackage{colortbl}   
\definecolor{iccvblue}{rgb}{0.21,0.49,0.74}
\definecolor{mycolor}{HTML}{D8ECD1}
\definecolor{dt}{gray}{0.7}
\definecolor{myblue}{RGB}{230,230,255}
\definecolor{lightgray}{gray}{0.91}
\definecolor{lightred}{RGB}{238,208,219}
\definecolor{promptstart}{RGB}{41, 128, 185}
\definecolor{promptend}{RGB}{52, 152, 219}
\definecolor{responsestart}{RGB}{46, 204, 113}
\definecolor{responseend}{RGB}{39, 174, 96}

\definecolor{ricebg}{RGB}{117, 255, 255}  
\definecolor{ricebg}{RGB}{173, 216, 230}  
\definecolor{ricebg}{RGB}{220, 255, 220}  
\definecolor{ricebg}{RGB}{230, 230, 250}  
\definecolor{ricebg}{RGB}{255, 229, 204}  
\definecolor{ricebg}{RGB}{255, 220, 220}  

\definecolor{ricebg}{RGB}{175, 238, 238}  
\definecolor{ricebg}{RGB}{152, 251, 152}  
\definecolor{ricebg}{RGB}{135, 206, 250}  
\definecolor{ricebg}{RGB}{255, 255, 204}  
\definecolor{ricebg}{RGB}{220, 208, 255}  

\definecolor{ricebg}{RGB}{236, 236, 236}  
\definecolor{ricebg}{RGB}{240, 248, 255}  
\definecolor{ricebg}{RGB}{245, 245, 220}  
\definecolor{ricebg}{RGB}{245, 255, 250}  
\definecolor{ricebg}{RGB}{240, 255, 240}  

\definecolor{ricebg}{RGB}{144, 238, 144}  
\definecolor{ricebg}{RGB}{188, 210, 238}  
\definecolor{negredd}{RGB}{238,208,219}  

\definecolor{header}{RGB}{255, 255, 255}
\definecolor{bordergray}{RGB}{80, 80, 80}

\definecolor{posgreen}{RGB}{0, 120, 0}
\definecolor{posgreen}{RGB}{0, 140, 0}
\definecolor{posgreen}{RGB}{0, 100, 0}
\definecolor{posgreen}{RGB}{20, 120, 20}
\definecolor{posgreen}{RGB}{60, 180, 60}

\usepackage{color}
\usepackage{colortbl}
\sethlcolor{yellow!30}

\usepackage{tcolorbox}
\usepackage{tikz}
\usetikzlibrary{shadows.blur}
\usepackage{fontawesome5}
\usepackage{caption}
\usepackage{xspace}
\usepackage[pagebackref,breaklinks,colorlinks,allcolors=iccvblue]{hyperref}

\newcolumntype{x}[1]{>{\centering\arraybackslash}p{#1pt}}
\newcolumntype{y}[1]{>{\raggedright\arraybackslash}p{#1pt}}
\newcolumntype{z}[1]{>{\raggedleft\arraybackslash}p{#1pt}}
\newcolumntype{^}{>{\currentrowstyle}}

\newcommand{\app}{\raise.17ex\hbox{$\scriptstyle\sim$}}
\newcommand{\repeatAmpersands}[1]{\ifnum#1>0 &\repeatAmpersands{\numexpr#1-1\relax}\fi}

\newcommand*{\belowrulesepcolor}[1]{%
  \noalign{%
    \kern-\belowrulesep
    \begingroup
      \color{#1}%
      \hrule height\belowrulesep
    \endgroup
  }%
}
\newcommand*{\aboverulesepcolor}[1]{%
  \noalign{%
    \begingroup
      \color{#1}%
      \hrule height\aboverulesep
    \endgroup
    \kern-\aboverulesep
  }%
}

\newtcolorbox{luxuryprompt}{
  enhanced,
  frame hidden,
  interior hidden,
  borderline west={4pt}{0pt}{promptstart},
  colback=blue!8!white,
  overlay={
    \begin{tcbclipinterior}
      \shade[left color=promptstart!20!white, right color=promptstart!5!white] 
        (interior.south west) rectangle (interior.north east);
      \draw[promptstart, line width=1.5pt] 
        ([xshift=12pt]frame.north west) -- ([xshift=-12pt]frame.north east);
      \draw[promptstart, line width=1.5pt] 
        ([xshift=12pt]frame.south west) -- ([xshift=-12pt]frame.south east);
      \node[inner sep=0pt] at ([xshift=9pt, yshift=-11pt]frame.north west) 
        {\includegraphics[width=22pt]{example-image-duck}};
    \end{tcbclipinterior}
  },
  sharp corners,
  boxsep=0pt,
  leftupper=18pt,
  rightupper=12pt,
  top=18pt,
  bottom=12pt,
  fontupper=\large\itshape\color{black!80!promptstart},
  toptitle=6pt,
  title={\normalfont\Large\bfseries\color{promptstart} \ding{229} User Prompt \ding{229}},
  coltitle=promptstart,
  attach title to upper,
  title style={center, scale=1.1},
  middle=1mm,
  shadow={10pt}{-3pt}{0pt}{black!10!white}
}

\title{Region-based Cluster Discrimination for Visual Representation Learning}

\author{Yin Xie†$^1$ \quad Kaicheng Yang†$^1$ \quad Xiang An†$^1$  \quad Kun Wu$^1$\quad Yongle Zhao$^1$\quad Weimo Deng$^1$\quad \\ Zimin Ran$^2$\quad Yumeng Wang$^1$\quad Ziyong Feng$^1$\quad Roy Miles $^3$\quad Ismail Elezi $^3$\quad Jiankang Deng $^4 \thanks{corresponding author, † equal contribution}$
\and
$^1$ DeepGlint \quad $^2$ University of Technology Sydney \quad \\ $^3$ 
Huawei London Research Center \quad $^4$ Imperial College London \\
{\tt\small \{yinxie,kaichengyang,xiangan\}@deepglint.com,jiankangdeng@gmail.com}
}

\begin{document}
\maketitle

\begin{abstract}
Learning visual representations is foundational for a broad spectrum of downstream tasks. Although recent vision-language contrastive models, such as CLIP and SigLIP, have achieved impressive zero-shot performance via large-scale vision-language alignment, their reliance on global representations constrains their effectiveness for dense prediction tasks, such as grounding, OCR, and segmentation. To address this gap, we introduce Region-Aware Cluster Discrimination (RICE), a novel method that enhances region-level visual and OCR capabilities. We first construct a billion-scale candidate region dataset and propose a Region Transformer layer to extract rich regional semantics. We further design a unified region cluster discrimination loss that jointly supports object and OCR learning within a single classification framework, enabling efficient and scalable distributed training on large-scale data. Extensive experiments show that RICE consistently outperforms previous methods on tasks, including segmentation, dense detection, and visual perception for Multimodal Large Language Models (MLLMs). The pre-trained models have been released at \url{https://github.com/deepglint/MVT}.
\end{abstract}

\section{Introduction}
The proliferation of large-scale image-text data on the Internet has driven remarkable advances in visual representation learning. Pioneering models such as CLIP~\cite{clip_icml}, DNF5B~\cite{DFN5B}, OpenCLIP~\cite{openclip}, EVA-CLIP~\cite{eva_clip}, AIM~\cite{aim}, ViTamin~\cite{chen2024vitamin}, and SigLIP~\cite{siglip} have effectively leveraged image-text alignment at scale to learn highly robust and generalizable visual features. As a result, these visual encoders have become foundational components in a wide range of computer vision applications, paving the way for more advanced MLLMs that integrate visual understanding with advanced language reasoning~\cite{li2024llavanext,liu2024llava_V1_5,li2024llavaov}.

\begin{figure*}[t]
    \centering
    \begin{subfigure}{0.225\textwidth}
        \includegraphics[width=\textwidth]{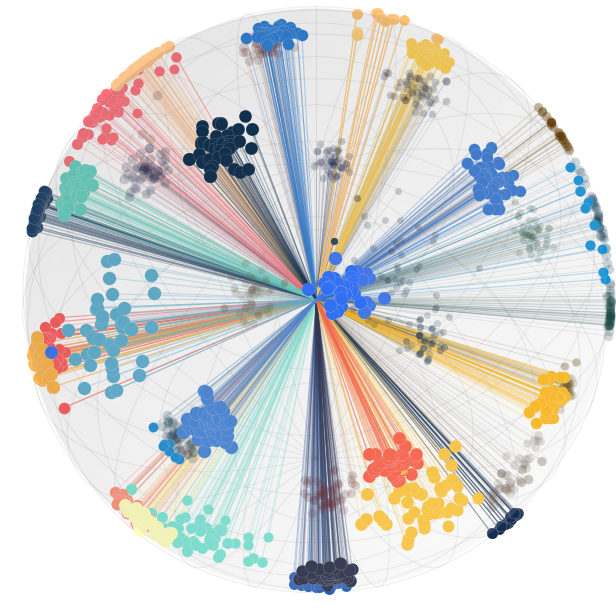}
        \caption{RICE (Ours)}
    \end{subfigure}%
    \hfill
    \begin{subfigure}{0.225\textwidth}
        \includegraphics[width=\textwidth]{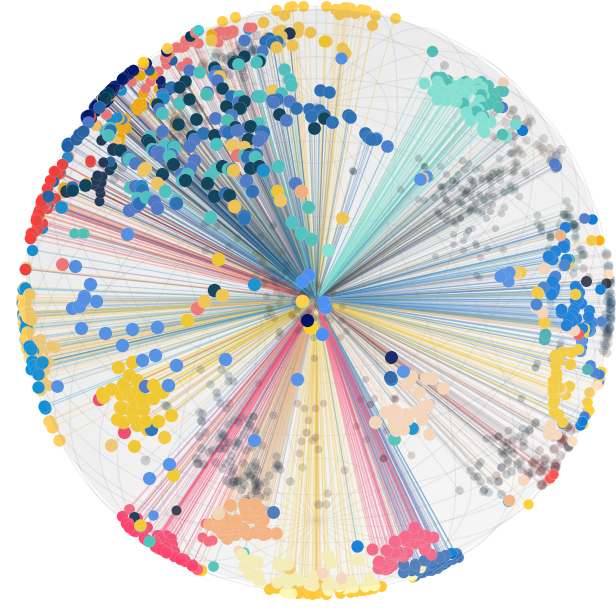}
        \caption{DINOv2}
    \end{subfigure}%
    \hfill
    \begin{subfigure}{0.225\textwidth}
        \includegraphics[width=\textwidth]{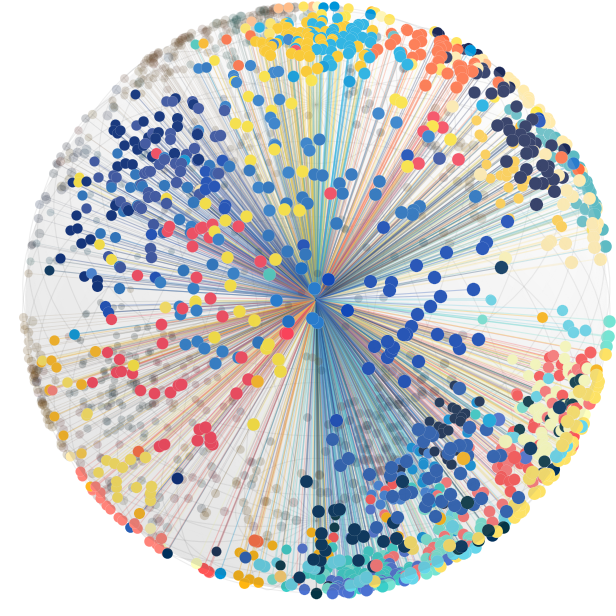}
        \caption{MLCD}
    \end{subfigure}%
    \hfill
    \begin{subfigure}{0.225\textwidth}
        \includegraphics[width=\textwidth]{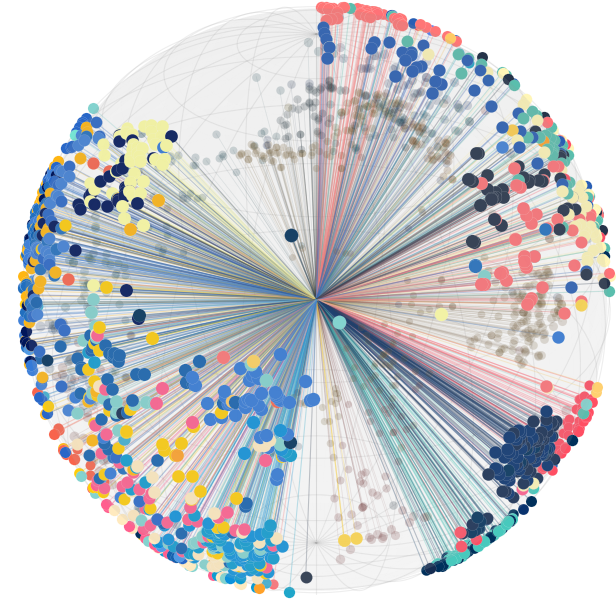}
        \caption{SigLIP}
    \end{subfigure}
    \vspace{-2mm}
    \caption{Visualization of object feature distributions from the COCO test dataset via t-SNE projection onto a spherical manifold.}
    \label{fig:coco_sphere}
    \vspace{-4mm}
\end{figure*}

However, these models generally struggle to capture the underlying semantic structure of the training data~(as shown in Fig.~\ref{fig:coco_sphere}). This shortcoming arises from instance-wise contrastive learning, which treats samples from different instances as negative pairs, regardless of their semantic similarity. Furthermore, in vision-language contrastive learning, pairs that involve Optical Character Recognition (OCR) often cause the visual encoder to focus primarily on spotting text. As a result, high-level visual semantics in the images are overshadowed, leading to reduced performance on downstream zero-shot tasks that require object-centric and scene-centric understanding~\cite{radenovic2023filtering}.

To overcome the limitations of instance discrimination, cluster discrimination methods such as DeepCluster~\cite{caron2018deep}, SeLa~\cite{asano2019self}, SwAV~\cite{caron2020unsupervised}, UNICOM~\cite{an2023unicom}, and MLCD~\cite{anxiang_2024_mlcd} have been proposed. These approaches jointly learn cluster assignments and image embeddings, encouraging similar instances to be grouped together, and thereby better capturing the underlying semantic structure of the data. However, most cluster discrimination methods rely on assigning one or multiple pseudo-labels to each image, which limits their ability to learn localized region-level representations.

Recent works have sought to achieve vision-language alignment at the local region level to support dense prediction tasks. RegionCLIP~\cite{regionclip} leverages the CLIP model to match image regions with template captions, pretraining the model to align region-text pairs in the feature space. CLIM~\cite{wu2023clim} creates mosaics by combining multiple images, treating each image as a pseudo region, and then extracts features for each region. A contrastive loss is applied to ensure that features of each pseudo region are similar to their corresponding text embeddings and dissimilar to others. Nevertheless, these approaches rely on the availability of descriptive text for regions, which introduces additional constraints and limits scalability to larger datasets.

In this work, we construct region-specific datasets tailored for both region-aware object and OCR tasks. We propose a unified learning framework based on cluster discrimination that jointly optimizes region-level object recognition and OCR capabilities. Within this framework, object regions are formulated as single-label classification tasks, while OCR regions are addressed as multi-label classification problems. Specifically, we first curate a large-scale candidate region dataset. For object regions, we utilize SAM~\cite{sam} to generate fine-grained mask regions for each image and extract CLIP features from these regions. Semantic labels are then assigned to each region using the $k$-means algorithm. For OCR regions, we employ PaddleOCR~\cite{du2020pp} to extract text data and generate corresponding OCR labels through tokenization.

To fully leverage the constructed region-specific datasets and enhance region-aware representation learning, we propose the RICE model, which integrates standard Transformer layers with novel Region Transformer layers. This architecture strengthens region-level representations while maintaining robust global feature learning. Additionally, we introduce an efficient region-cluster classification loss that unifies object and OCR learning within a single framework, enabling scalable distributed training on large-scale datasets. Extensive experiments show that RICE consistently outperforms previous methods across a range of downstream tasks, including segmentation, dense detection, and visual perception for MLLMs. In summary, our main \textbf{contributions} are as follows:
\begin{itemize}
\item We \textbf{propose} the RICE model for region-aware representation learning, introducing the Region Transformer layer as a core module for extracting rich regional semantic information.
\item We \textbf{design} an efficient region-based cluster discrimination loss that unifies object and OCR learning within a single classification framework, enabling scalable training on large-scale datasets.
\item We \textbf{conduct} extensive experiments, demonstrating the effectiveness of our approach across multiple downstream tasks, including segmentation, dense detection, and visual perception for MLLMs.
\end{itemize}

\section{Related Work}
\noindent{\bf Image Representation Learning.}
Existing image representation learning methods can be categorized into two approaches: (1) instance-based discrimination and (2) cluster-based discrimination. Instance-based methods, exemplified by CLIP~\cite{clip_icml,yang2023alip,gu2024rwkvclip,yang2025clipcid} and SigLIP~\cite{siglip}, leverage a simple yet effective strategy that utilizes diverse label forms across different granularities for individual images. DINOv2~\cite{oquab2024dinov2learningrobustvisual} introduces a self-supervised framework that does not require labels, achieving state-of-the-art results across a wide range of vision tasks. UNIT~\cite{unit_hangxu} integrates lightweight language and vision decoders, enabling robust text prediction while mitigating catastrophic forgetting in visual encoding. However, a key limitation of instance-wise contrastive learning is that it treats all different instances as negative pairs, which restricts its capacity to capture richer semantic relationships in the data. To address this limitation, cluster discrimination approaches such as UNICOM~\cite{an2023unicom} and MLCD~\cite{anxiang_2024_mlcd} have been proposed. These methods group multiple instances into clusters, drawing visually similar samples closer together and thus better capturing the underlying semantic structure. Nonetheless, most cluster discrimination methods assign one or more pseudo-labels to each image, which can hinder their ability to learn effective region-level representations.

\noindent{\bf Region Representation Learning.}
Early object detection methods~\cite{redmon2016you,ren2015faster} typically rely on manually labeled regions to train their models. To alleviate the substantial cost of manual annotation, PreDet~\cite{ramanathan2021predet} extends self-supervised learning to region representation by enforcing similar embeddings for highly overlapping bounding boxes across different views of the same image, while contrasting them with non-overlapping regions. RegionCLIP~\cite{regionclip} leverages the CLIP model to align image regions with template captions, pretraining the model to map region-text pairs into a shared feature space. GLIP~\cite{li2021grounded} exploits large-scale image-text pairs by generating grounding boxes through self-training, thereby producing semantically rich representations. CLIM~\cite{wu2023clim} constructs mosaics from multiple images, treating each image as a pseudo-region to facilitate training.
Distinct from these approaches, we propose leveraging region cluster centers as supervision signals. This strategy enhances semantic richness and facilitates scalable training by removing the reliance on descriptive textual annotations.

\begin{figure*}[t!]
\centering
\includegraphics[width=0.98\linewidth]{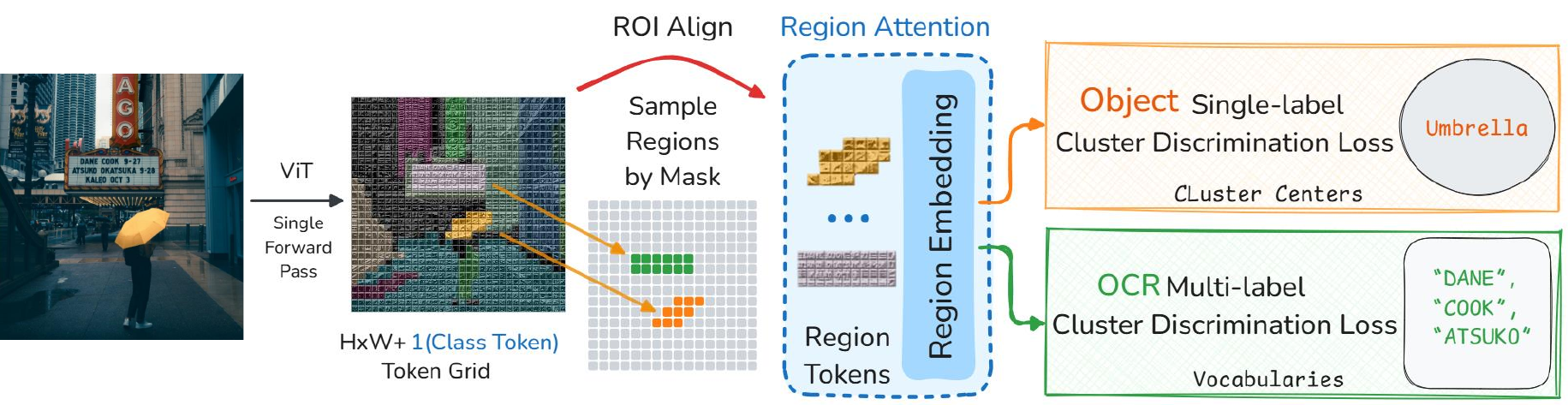}
\vspace{-2mm}
\caption{Overview of our unified semantic region understanding framework. Our approach efficiently processes diverse semantic regions within the image using a single forward pass. The model jointly captures both general visual semantics (objects) and OCR semantics (texts), seamlessly integrating them into a unified representation.}
\label{fig:masks}
\vspace{-4mm}
\end{figure*}

\noindent{\bf Multimodal Large Language Model.}
Motivated by the reasoning capabilities of Large Language Models (LLMs)~\cite{llama2, vicuna2023, qwen1, qwen2}, recent research has explored transferring these abilities to the vision domain through the development of multimodal LLMs~\cite{tong2024cambrian,Qwen-VL}. Approaches such as Flamingo~\cite{alayrac2022flamingo} and BLIP~\cite{li2022blip,li2023blip} connect pre-trained vision encoders with frozen language models to enable multimodal understanding, while LLaVA~\cite{liu2024llava_V1_5,li2024llavaov} achieves efficient vision-language alignment via instruction tuning. Recent advances in region-aware multimodal LLMs, such as MiniGPT-4~\cite{zhu2023minigpt} and LLaVA-G~\cite{llava_ground_2024_eccv}, incorporate region information in textual form, relying primarily on the language decoder for positional interpretation. LISA~\cite{lai2024lisa} pioneered reasoning-based segmentation methods, and GLaMM~\cite{rasheed2024glamm} introduced new datasets and techniques for region-level captioning and segmentation. Concurrently, works such as~\cite{wujiannan_visionLLMv2,wei2024lasagna,Xia2023GSVAGS,PixelLM_2024_CVPR,yuan2025sa2va} investigate integrating referring segmentation with conversational interfaces through instruction tuning. However, performance on complex region understanding tasks remains limited by the region representation capabilities of the image encoder.

\section{Methodology}
In this paper, we enhance the semantic object perception and OCR capabilities of visual representation models. To this end, we first construct region-aware datasets (Sec.~\ref{regiondata}), providing semantically rich supervision. We then introduce a Region Transformer layer (Sec.~\ref{ricemodel}) to effectively leverage these data. Finally, we employ a region cluster discrimination loss (Sec.~\ref{sec:region_contrastive_learning}) that unifies general object recognition and OCR within a single classification framework.

\subsection{Region Data Curation}
\label{regiondata}

\noindent{\bf Region-Aware Object Data Curation.}
To construct region-aware object data, we first sample images from the LAION2B~\cite{schuhmann2022laion}, COYO700M~\cite{kakaobrain2022coyo-700m}, and SAM1B~\cite{sam} datasets, ensuring each image has a minimum edge length of 336 pixels. For LAION2B and COYO700M, we apply the SAM model to generate fine-grained mask regions, while for SAM1B, we retain the original annotated regions. We further filter candidate bounding boxes to those with a minimum edge length of 128 pixels, resulting in a dataset comprising 400 million images and 2 billion candidate regions. Following UNICOM~\cite{an2023unicom}, we extract features from these regions using the CLIP model and apply the $k$-means algorithm to cluster them into one million semantic centers, $\mathcal{C}={\mathbf{c}_1, \mathbf{c}_2, ..., \mathbf{c}_K}$.
The semantic label for each region is then assigned by nearest-neighbor matching to the cluster center:
\begin{equation}
\mathbf{y}^{object}_{i,j} = \underset{k \in [1,K]}{\arg\min} \|\mathbf{f}_{i,j} - \mathbf{c}_k\|_2,
\end{equation}
where $\mathbf{f}_{i,j} = \mathcal{F}_{\text{CLIP}}(\mathbf{r}_{i,j})$ denotes the CLIP feature embedding of region $\mathbf{r}_{i,j}$, which is the $j$-th semantic region in the $i$-th image. 
Overall, the clustering and label assignment step is relatively efficient. We implemented and parallelized this process using hierarchical k-means with the Faiss GPU library~\cite{douze2025faisslibrary}, completing the task in approximately 10 hours on 64 graphics cards.

\noindent{\bf Region-Aware OCR Data Curation.}
For OCR data, we use PaddleOCR~\cite{du2020pp} to extract text information from the LAION2B and COYO700M datasets, retaining only entries with a confidence score above 0.7. This yields a dataset of 50 million images and 400 million candidate regions. We then tokenize the extracted text for each region to obtain the corresponding OCR labels, $\mathbf{y}^{ocr}_{i,j} = tokenizer(\mathbf{t}_{i,j})$, where $\mathbf{t}_{i,j}$ denotes the text associated with region $\mathbf{r}_{i,j}$.
The final supervision signal combines semantic object cluster labels ($\mathbf{y}^{object}$) and OCR-derived labels ($\mathbf{y}^{ocr}$), depending on the selected region. The integration of these labels into the region-aware cluster discrimination framework is detailed in Sec.~\ref{sec:region_contrastive_learning}.

\subsection{Region-aware Representation}
\label{ricemodel}

Our framework is designed to achieve region-aware representation in a single inference pass. Unlike current visual encoders such as CLIP~\cite{clip_icml} and DINOv2~\cite{oquab2024dinov2learningrobustvisual}, our approach integrates standard Transformer layers for global semantic understanding with specialized Region Transformer layers for extracting regional semantics.

\noindent{\bf Region Sampling.}
To handle the varying number and size of regions in each image, we employ a balanced sampling strategy that standardizes the number of regions to $N$ for efficient data loading. If the original number of regions $|\mathcal{R}_i|$ exceeds $N$, we randomly sample $N$ regions from the index set $\mathcal{I}_{\mathcal{R}_i}$. Conversely, if $|\mathcal{R}_i| < N$, we retain all existing regions and augment the set by randomly resampling from $\mathcal{I}_{\mathcal{R}_i}$ until $N$ regions are reached. This adaptive strategy ensures a consistent computational cost while preserving all available information when the region count is low.

\begin{figure}[t!]
\centering
\includegraphics[width=0.95\linewidth]{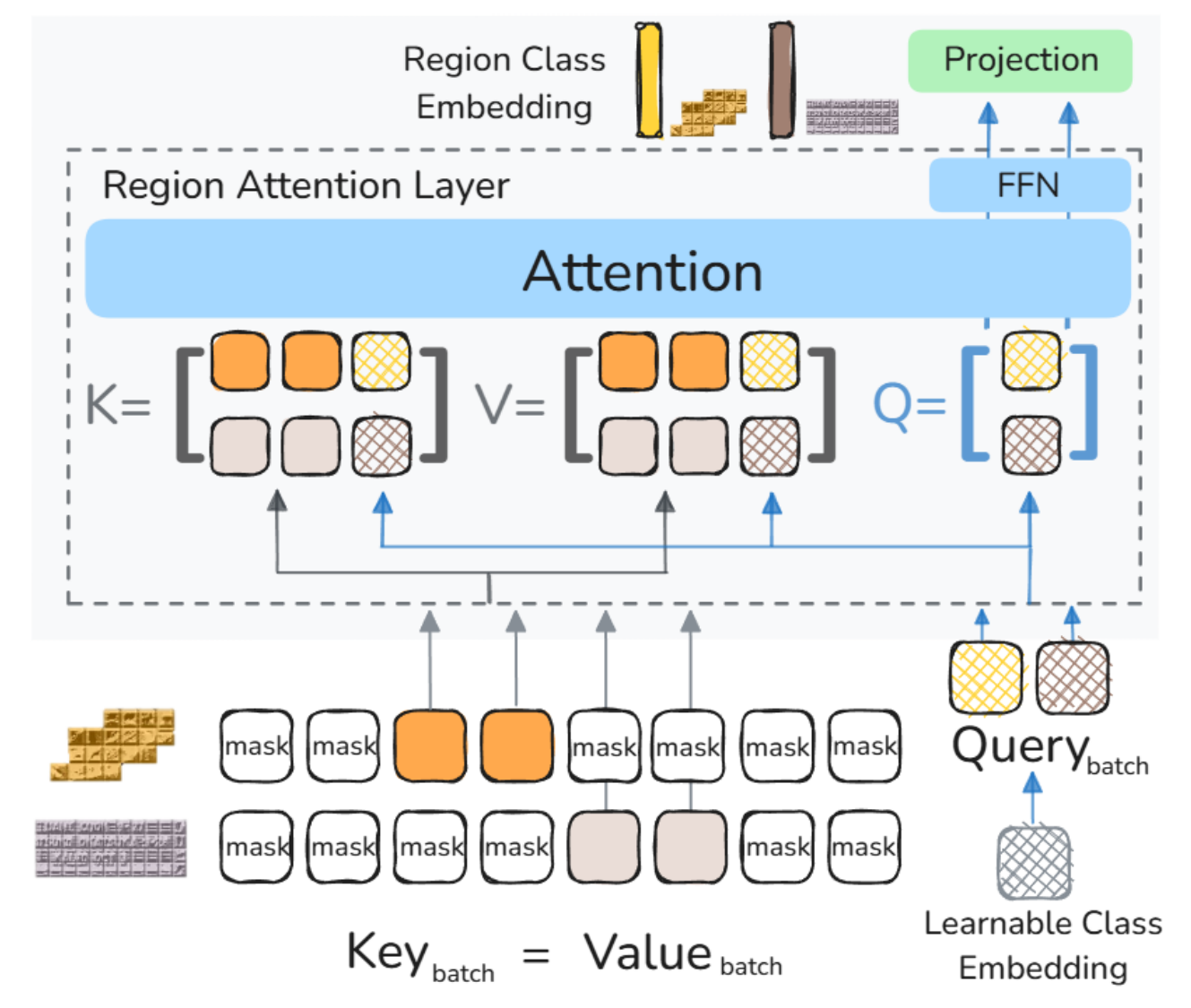}
\vspace{-2mm}
\caption{The region attention module processes batches of size-varying regions by leveraging a region-specific visibility mask and produces fixed-length region class embeddings. This approach enables efficient scaling of training while preserving representation fidelity.}
\vspace{-4mm}
\label{fig:attention}
\end{figure}

\noindent{\bf Region Attention Layer.}
We adopt a sampling strategy that ensures a consistent number of regions per image. However, since regions vary in spatial size, the number of tokens per region remains inconsistent, complicating batch processing and hindering training scalability. To overcome this, we introduce a region attention layer as illustrated in Fig.~\ref{fig:attention}. The proposed region attention layer applies a region-specific visibility mask to restrict attention to tokens within each designated region. This masking mechanism allows regions of different sizes to be efficiently grouped within the same batch, thereby facilitating scalable training.

Given an image $\mathbf{x} \in \mathbb{R}^{H \times W \times C}$, we first apply a 2D convolutional layer to compute patch tokens $\mathbf{z} = \left\{\boldsymbol{z}_{\text{cls}}, \boldsymbol{z}_1, \boldsymbol{z}_2, \ldots, \boldsymbol{z}_{N}\right\} \in \mathbb{R}^{(N+1) \times D}$, where $N = \frac{H}{p} \times \frac{W}{p}$ and $p$ is the patch size. These tokens are then fed into Vision Transformer layers to produce global semantic embeddings $\mathcal{E} = \left\{\boldsymbol{e}_{\text{cls}}, \boldsymbol{e}_1, \boldsymbol{e}_2, \ldots, \boldsymbol{e}_{N}\right\}$. 
Subsequently, we input these embeddings into the proposed Region Transformer layers to extract regional semantics. By employing standard global attention in the initial layers and region-aware attention in the later layers, our approach effectively balances the additional training overhead introduced by region-aware attention (whose batch size scales with $N$), while still preserving global contextual information.

As illustrated in Fig.~\ref{fig:attention}, our proposed Region Attention Layer differs from the standard Vision Transformer by introducing a novel mask-guided region attention mechanism. To enable efficient parallel processing of all regions, we employ a region-specific visibility mask $\mathcal{M}$, which assigns a value of $0$ to tokens within the region and $-\infty$ to those outside the region. This visibility mask $\mathcal{M}$ enforces region-constrained attention and facilitates batching across regions of varying sizes. The region attention operation is mathematically defined as:
\begin{equation}
\mathcal{R}_{\text{batch}} = \sigma\left(\frac{\boldsymbol{Q}_{\text{batch}}\boldsymbol{K}_{\text{batch}}^{\mathsf{T}}}{\sqrt{d_k}} + \mathcal{M} \right)\boldsymbol{V}_{\text{batch}},
\end{equation}
where $\sigma(\cdot)$ denotes the softmax normalization function. This formulation enables the extraction of multiple region class embeddings in a single forward pass, optimizing computational efficiency while preserving representation fidelity.
In terms of dimensions, the Query matrix $\boldsymbol{Q}_{\text{batch}}$ has the shape $(L, B, D)$, where $L$ is the number of regions processed, $B$ is the batch size, and $D$ is the embedding dimension. The Key and Value matrices, $\boldsymbol{K}_{\text{batch}}$ and $\boldsymbol{V}_{\text{batch}}$, include all tokens of the original image. After the attention operation, the output $\mathcal{R}_{\text{batch}}$ retains the same dimensions as the Query, yielding a $(L, B, D)$ tensor that provides fixed-length representations for all regions.

\subsection{Region Cluster Discrimination}
\label{sec:region_contrastive_learning}

After obtaining the embedding representation for each region, we introduce two specialized loss functions: (1) Object Region Loss for general visual recognition, and (2) OCR Region Loss for text recognition.

\noindent{\bf Object Region Loss.} In the Object Region Loss, each region is associated with a single positive class center selected from a pool of one million class centers. During training, the embeddings are encouraged to move closer to their corresponding positive class center while simultaneously being pushed away from a randomly sampled subset of negative class centers~\cite{an2023unicom}. The Object Region Loss is mathematically defined as follows:
\begin{align}
\label{eqn:loss_gen}
\mathcal{L}_\mathrm{object} = &\log(1 + \exp({-sim(\hat{\mathbf{y}}_{i,j}, \mathbf{y}^{object}_{i,j}}))
\\[0.5em]
&+ \log(1 + \sum_{j \in \Omega^{object}_n} \exp({sim(\hat{\mathbf{y}}_{i,j}, \mathbf{y}^{object}_{i,j}})),
\nonumber
\end{align}
where $sim(\cdot)$ denotes the similarity function between two region features, and $\hat{\mathbf{y}}_{i,j}$ represents the feature embedding for the $j$-th region in the $i$-th image. Each region has a single positive class center, denoted as $\mathbf{y}^{object}_{i,j}$, while multiple negative class centers are randomly sampled from the set of negative centroids $\Omega_n^{object}$. 

\noindent{\bf OCR Region Loss.} In the OCR Region Loss, positive classes are directly obtained from the token embeddings of the text within each OCR region. Each token embedding serves as a positive class center, resulting in multiple positive classes per OCR region. Consequently, the OCR Region Loss is designed to accommodate the intrinsic correspondence between text regions and multiple positive class centers ($M$)~\cite{anxiang_2024_mlcd}. The OCR loss is defined as follows:
\begin{align}
\label{eqn:loss_ocr}
\mathcal{L}_\mathrm{ocr} = &\log(1 + \sum_{j \in \Omega^{ocr}_p} \exp({-sim(\hat{\mathbf{y}}_{i,j}, \mathbf{y}^{ocr}_{i,j}})) 
\\
&+ \log(1 + \sum_{j \in \Omega^{ocr}_n} \exp({sim(\hat{\mathbf{y}}_{i,j}, \mathbf{y}^{ocr}_{i,j}})),
\nonumber
\end{align}
where the negative labels are sampled from all other token embeddings, denoted as $\Omega_n^{ocr}$. As previously noted, token embeddings computed within the same OCR region are selected as positives, denoted by $\Omega_p^{ocr}$.

When conflicting classes are present, they can generate incorrect gradient signals that degrade model performance. To address this, we employ the random selection strategy~\cite{An2022PartialFC} to efficiently construct a subset of negative class centers. Specifically, negative classes are uniformly sampled as $\Omega_n = \Omega_{\text{uniform}}^{k}({1,2,\ldots,K} \setminus \Omega_p)$, where $k = \lfloor K \times \rho \rfloor$ and $\rho \in (0,1]$ controls the sampling density. This strategy offers three main benefits: it improves computational efficiency by reducing overhead, lowers the chance of including semantically similar classes as negatives thus reducing conflicting gradients, and promotes more stable model convergence. Experiments show that an appropriate sampling ratio $\rho$ significantly enhances performance.

\section{Experiments}

\subsection{Implementation Details}
\noindent{\bf Pretraining Setup.} We train our RICE models on LAION-2B~\cite{schuhmann2021laion}, SAM1B~\cite{sam}, and COYO-700M~\cite{kakaobrain2022coyo-700m}, processing a total of 13 billion samples during the initial pretraining stage. Training is conducted with a global batch size of 32K distributed across 64 graphics cards. We utilize the AdamW~\cite{loshchilov2018decoupled} optimizer with a learning rate of 0.001 and a decoupled weight decay coefficient of 0.2. In this paper, we employ ViT-L/14 and ViT-B/16 architectures. For ViT-L/14, we initially train at a resolution of $224 \times 224$, followed by fine-tuning at $336 \times 336$, $378 \times 378$, and $560 \times 560$. For ViT-B/16, training starts at $224 \times 224$ and proceeds to $512 \times 512$ for fine-tuning. At higher resolutions, the learning rate is reduced by an order of magnitude, and 1 billion samples are used for fine-tuning. Throughout all experiments, the number of classes ($K$) is fixed at one million. Following the margin-based classification approach~\citep{deng2019arcface}, we apply L2 normalization to both the feature vectors and class centers, and set a margin of $m=0.3$ for positive classes with a scale parameter of 64. The ratio of sampled negative class centers ($\rho$) is fixed at 0.1, as in previous work~\cite{an2023unicom,anxiang_2024_mlcd}.

\noindent{\bf Multimodal Setup.} For our multimodal large language model evaluations, we adopt the LLaVA-NeXT~\cite{li2024llavanext,liu2024llavanext} framework to ensure experimental consistency. All training procedures strictly follow the original LLaVA-NeXT implementation, using the same pretraining datasets and instruction-tuning data. We utilize Qwen2.5-7B~\cite{qwen2.5} as the language model backbone to avoid hyperparameter biases that could favor the OpenAI CLIP model in the original setup. This controlled experimental design enables a fair comparison when evaluating the performance of our vision encoders within multimodal systems.

\noindent{\bf Referring Image Segmentation Setup.}
For our referring image segmentation evaluation, we integrate our vision encoders into LLaVA-NeXT using Qwen2.5-7B as the language model backbone. Following LISA~\cite{lai2024lisa}, we adopt their two-stage training approach, vision-language alignment followed by MLLM-Decoder training with SAM integration. We also employ the same training data mixture of semantic and referring segmentation datasets. Additionally, we incorporate the LLaVA-NeXT 740K instruction dataset~\cite{li2024llavanext}. For segmentation tasks, we introduce a specialized \textcolor{responsestart!80!black}{[SEG]} token, whose embedding is transformed via an MLP adapter to generate SAM prompts.

\begin{table*}[!t]
    \centering
    \fontsize{6.8pt}{6.0pt}\selectfont
    \setlength\tabcolsep{2.95pt}
    \renewcommand{\arraystretch}{1.55}

    \begin{tabular}{lll|*{7}{c}|c|*{7}{c}|c}
        \toprule
        \rowcolor{header!8}
        \multicolumn{3}{c|}{\textbf{Model Configuration}} & 
        \multicolumn{8}{c|}{\textbf{OCR \& Document Understanding}} & 
        \multicolumn{8}{c}{\textbf{General Vision Understanding}} \\
        
        \midrule
        \textbf{Method} & \textbf{Vision Tower} & \textbf{LLM} & 
        \rotatebox{90}{InfoVQA\cite{vlm_infovqa}} & 
        \rotatebox{90}{DocVQA\cite{vlm_docvqa}} & 
        \rotatebox{90}{ChartQA\cite{vlm_chartqa}} & 
        \rotatebox{90}{TextVQA\cite{vlm_textvqa}} & 
        \rotatebox{90}{OCRBench\cite{vlm_ocrbench}} & 
        \rotatebox{90}{OCRBenchV2\cite{vlm_ocrbench}} & 
        \rotatebox{90}{LiveXivVQA\cite{vlm_livexiv}} & 
        \rotatebox{90}{\textbf{OCR Avg}} & 
        \rotatebox{90}{AI2D\cite{vlm_ai2d}} & 
        \rotatebox{90}{MMB$^\text{EN}$\cite{vlm_mmbench}} & 
        \rotatebox{90}{MME$^\text{Cog}$\cite{vlm_mme}} & 
        \rotatebox{90}{MME$^\text{Per}$\cite{vlm_mme}} & 
        \rotatebox{90}{POPE\cite{vlm_pope}} & 
        \rotatebox{90}{RealworldQA\cite{vlm_realworldqa}} & 
        \rotatebox{90}{MMStar\cite{vlm_mmstar}} & 
        \rotatebox{90}{\textbf{Other Avg}} \\
        \midrule
        \multicolumn{19}{l}{\textbf{LLaVA-NeXT Framework}} \\
        \hline
        CLIP~\cite{clip_icml} & ViT-L-14-336px & Qwen2.5-7B & 38.88 & 75.21 & 66.52 & 62.47 & 525 & 22.95 & 47.35 & 52.27 & 73.15 & 74.57 & 384 & 1512 & 88.83 & 63.66 & 48.98 & 69.83 \\
        MLCD~\cite{anxiang_2024_mlcd} & ViT-L-14-336px & Qwen2.5-7B & 43.48 & 76.46 & 67.84 & 61.69 & 531 & 23.98 & 48.43 & 53.57 & 76.98 & 76.37 & 433 & 1598 & 88.69 & 61.05 & 50.98 & 72.31 \\
        AIMv2~\cite{fini2025multimodal} & ViT-L-14-336px & Qwen2.5-7B & 35.44 & 77.19 & 72.72 & 65.85 & 572 & 23.92 & 47.28 & 54.23 & 75.36 & 78.61 & 386 & 1500 & 88.41 & 62.22 & 50.16 & 70.58 \\
        \rowcolor{ricebg}RICE & ViT-L-14-336px & Qwen2.5-7B & 45.23 & 79.19 & 72.34 & 65.89 & 575 & 24.14 & 48.94 & 56.18 & 77.88 & 76.55 & 437 & 1613 & 88.49 & 63.05 & 51.75 & 73.03 \\
        \multicolumn{3}{l|}{↑ RICE-336px vs AIMv2} & \textcolor{posgreen}{+9.79} & \textcolor{posgreen}{+2.00} & \textcolor{negredd}{-0.38} & \textcolor{posgreen}{+0.04} & \textcolor{posgreen}{+3} & \textcolor{posgreen}{+0.22} & \textcolor{posgreen}{+1.66} & \textcolor{posgreen}{+1.95} & \textcolor{posgreen}{+2.52} & \textcolor{negredd}{-2.06} & \textcolor{posgreen}{+51} & \textcolor{posgreen}{+113} & \textcolor{posgreen}{+0.08} & \textcolor{posgreen}{+0.83} & \textcolor{posgreen}{+1.59} & \textcolor{posgreen}{+2.45} \\
        \hline
        DFN5B~\cite{DFN5B} & ViT-H-14-378px & Qwen2.5-7B & 38.59 & 70.87 & 64.36 & 59.40 & 473 & 21.88 & 46.17 & 49.80 & 73.51 & 73.37 & 366 & 1537 & 88.62 & 59.87 & 49.14 & 68.91 \\
        UNIT~\cite{unit_hangxu} & ViT-H-14-378px & Qwen2.5-7B & 42.83 & 77.21 & 69.94 & 65.41 & 559 & 22.94 & 46.43 & 54.38 & 74.24 & 73.02 & 354 & 1554 & 88.33 & 61.00 & 46.46 & 68.54 \\
        SigLIP~\cite{siglip} & ViT-SO400M-14-384px & Qwen2.5-7B & 41.38 & 76.71 & 69.28 & 64.74 & 554 & 23.97 & 48.42 & 54.27 & 76.17 & 76.98 & 369 & 1597 & 88.83 & 63.66 & 47.32 & 70.62 \\
        SigLIPv2~\cite{siglipv2} & ViT-SO400M-14-384px & Qwen2.5-7B & 43.66 & 79.14 & 70.16 & 66.23 & 587 & 25.35 & 48.59 & 55.98 & 76.98 & 77.14 & 373 & 1607 & 89.26 & 63.36 & 52.77 & 71.70 \\
        \rowcolor{ricebg}RICE & ViT-L-14-378px & Qwen2.5-7B & 48.05 & 82.59 & 75.08 & 66.24 & 588 & 25.84 & 49.54 & 58.02 & 76.54 & 77.60 & 433 & 1580 & 89.14 & 62.88 & 51.20 & 72.63 \\
        \multicolumn{3}{l|}{↑ RICE-378px vs SigLIPv2} & \textcolor{posgreen}{+4.39} & \textcolor{posgreen}{+3.45} & \textcolor{posgreen}{+4.92} & \textcolor{posgreen}{+0.01} & \textcolor{posgreen}{+1} & \textcolor{posgreen}{+0.49} & \textcolor{posgreen}{+0.95} & \textcolor{posgreen}{+2.04} & \textcolor{negredd}{-0.44} & \textcolor{posgreen}{+0.46} & \textcolor{posgreen}{+60} & \textcolor{negredd}{-27} & \textcolor{negredd}{-0.12} & \textcolor{negredd}{-0.48} & \textcolor{negredd}{-1.57} & \textcolor{posgreen}{+0.93} \\
        \hline
        SigLIPv2~\cite{siglipv2} & ViT-SO400M-16-560px & Qwen2.5-7B & 50.23  & 86.20  & 77.40 & 70.23 & 627 & 26.47 & 52.94 & 60.88 & 76.98 & 76.46 & 428 & 1597 & 89.26 & 68.24 & 53.06 & 73.60 \\
        Qwen2.5-ViT~\cite{qwen2_5_vl} & ViT-H-14-560px & Qwen2.5-7B & 55.91 & 85.83 & 78.84 & 73.70 & 662 & 26.84 & 53.44 & 62.97 & 78.76 & 78.44 & 496 & 1616 & 88.64 & 64.18 & 54.98 & 75.50 \\
        \rowcolor{ricebg}RICE & ViT-L-14-560px & Qwen2.5-7B & 53.15 & 87.38 & 78.08 & 68.96 & 607 & 26.14 & 53.03 & 61.06 & 76.88 & 78.61 & 450 & 1585 & 88.94 & 65.10 & 50.47 & 73.46 \\
        \multicolumn{3}{l|}{↑ RICE-560px vs SigLIPv2-560px} & \textcolor{posgreen}{+2.92 } & \textcolor{posgreen}{+1.18 } & \textcolor{posgreen}{+0.68} & \textcolor{negredd}{-1.27} & \textcolor{negredd}{-20} & \textcolor{negredd}{-0.33} & \textcolor{posgreen}{+0.09} & \textcolor{posgreen}{+0.18} & \textcolor{negredd}{-0.10} & \textcolor{posgreen}{+2.15} & \textcolor{posgreen}{+22} & \textcolor{negredd}{-12} & \textcolor{negredd}{-0.32} & \textcolor{negredd}{-3.14} & \textcolor{negredd}{-2.59} & \textcolor{negredd}{-0.14} \\
        \hline
        \multicolumn{19}{l}{\textbf{LLaVA-OneVision Framework}} \\
        \hline
        SigLIP~\cite{siglip} & ViT-SO400M-14-384px & Qwen2-7B & 68.80 & 87.50 & 80.00 & 78.50 & 697 & 27.10 & 58.44 & 67.15 & 81.40 & 80.80 & 418 & 1580 & 87.20 & 66.30 & 61.70 & 75.15 \\
        \rowcolor{ricebg}RICE & ViT-L-14-378px & Qwen2-7B & 73.57 & 91.80 & 82.71 & 79.59 & 749 & 32.05 & 60.96 & 70.80 & 84.57 & 81.75 & 564 & 1601 & 88.83 & 68.94 & 63.88 & 80.29 \\
        \multicolumn{3}{l|}{↑ RICE vs SigLIP (OneVision)} & \textcolor{posgreen}{+4.77} & \textcolor{posgreen}{+4.30} & \textcolor{posgreen}{+2.71} & \textcolor{posgreen}{+1.09} & \textcolor{posgreen}{+52} & \textcolor{posgreen}{+4.95} & \textcolor{posgreen}{+2.52} & \textcolor{posgreen}{+3.65} & \textcolor{posgreen}{+3.17} & \textcolor{posgreen}{+0.95} & \textcolor{posgreen}{+146} & \textcolor{posgreen}{+21} & \textcolor{posgreen}{+1.63} & \textcolor{posgreen}{+2.64} & \textcolor{posgreen}{+2.18} & \textcolor{posgreen}{+5.14} \\
        \bottomrule
    \end{tabular}
    \vspace{-2mm}
    \caption{
    Comprehensive performance comparison of RICE with state-of-the-art vision encoders. For all experiments within the LLaVA-NeXT framework, we adopt a high-resolution tiling strategy: each input image is divided into a $2\times2+1$ grid of crops, where each crop matches the pre-training resolution of the backbone model (\eg, 336px, 378px, or 560px). The scores for OCRBench, MME$^\text{Cog}$, and MME$^\text{Per}$ are normalized (divided by 10, 8, and 20, respectively) to scale all results to a 0--100 range.
    }
    \label{tab:comparison_in_llava}
    \vspace{-4mm}
\end{table*}

\subsection{RICE as a Vision Encoder for MLLMs}
\noindent{\bf Performance of Multimodal Understanding.} 
Based on Tab.~\ref{tab:comparison_in_llava}, we comprehensively evaluate the RICE model across multiple vision backbones within leading multimodal large language model frameworks. At a resolution of 336px, RICE achieves substantial performance gains over the widely used CLIP model and consistently outperforms more complex models with higher input resolutions, such as SigLIP and DFN5B. Notably, RICE delivers significant improvements on OCR-related tasks: on OCRBench, RICE surpasses CLIP-336px by +50 points and SigLIP-384px by +34 points, and on DocVQA, RICE achieves improvements of +3.98\%, +5.68\%, and +4.30\% over the respective baselines. Similar trends are observed for InfoVQA and TextVQA. At higher resolutions (560px), RICE maintains its lead, outperforming SigLIPv2-560px by +2.92\% on InfoVQA and +1.18\% on DocVQA. Remarkably, RICE-560px attains a DocVQA score of 87.38\%, exceeding the performance of Qwen2.5-VL’s specialized backbone (85.83\%). Furthermore, across both LLaVA-NeXT and LLaVA-OneVision frameworks, RICE outperforms CLIP and SigLIP on most benchmarks, with particularly strong gains in perception tasks. Collectively, these results establish RICE as a robust and versatile vision encoder, achieving substantial improvements on OCR-centric benchmarks, maintaining strong general vision performance, and delivering results competitive with specialized models trained on extensive datasets.

\begin{table*}[t!]
\centering
\resizebox{0.87\linewidth}{!}{
    \begin{tabular}{lllcccccccc@{}}
        \toprule
        \multirow{2}{*}{VisionTower}&\multirow{2}{*}{LLM}&\multirow{2}{*}{Method}&\multicolumn{3}{c}{refCOCO~\cite{kazemzadeh2014referitgame}}&\multicolumn{3}{c}{refCOCO+~\cite{kazemzadeh2014referitgame}}&\multicolumn{2}{c}{refCOCOg~\cite{refcocog}}\\
        &&&val&testA&testB&val&testA&testB&val&test\\\midrule
        \multicolumn{11}{l}{\textit{\textbf{Previous Methods}}} \\\midrule
        CLIP&Vicuna-7B~\cite{vicuna2023}&GLaMM~\cite{rasheed2024glamm}&79.5&83.2&76.9&72.6&78.7&64.6&74.2&74.9\\
        CLIP&Vicuna-7B~\cite{vicuna2023}&VisionLLMv2~\cite{wujiannan_visionLLMv2}&79.2&82.3&77.0&68.9&75.8&61.8&73.3&74.8\\
        CLIP&Vicuna-7B~\cite{vicuna2023}&LLaVA-G-7B~\cite{llava_ground_2024_eccv}&77.1&-&-&68.8&-&-&71.5&-\\
        CLIP&LLaMA2-13B~\cite{llama2}&GSVA~\cite{Xia2023GSVAGS}&79.2&81.7&77.1&70.3&73.8&63.6&75.7&77.0\\
        CLIP&LLaMA2-13B~\cite{llama2}&PixelLM-7B~\cite{PixelLM_2024_CVPR}&73.0&-&-&66.3&-&-&69.3&-\\
        ConvNext-L~\cite{convnext,openclip}& InternLM2-7B~\cite{internlm2}&OMG-LLaVA\cite{NEURIPS2024_83eb86be}&77.2&79.8&74.1&68.7&73.0&61.6&71.7&71.9\\
        InternViT2.5-300M~\cite{chen2024expanding}&InternLM2.5-7B~\cite{chen2024expanding}&Sa2VA~\cite{yuan2025sa2va}&81.6&-&-&76.2&-&-&78.7&-\\
        InternViT2.5-6B~\cite{chen2024expanding}&InternLM2.5-20B~\cite{chen2024expanding}&Sa2VA~\cite{yuan2025sa2va}&82.5&-&-&78.8&-&-&79.7&-\\
        \midrule
        \multicolumn{11}{l}{\textit{\textbf{LLaVA-1.5 Framework}}} \\
        \midrule
        CLIP&Vicuna-7B~\cite{vicuna2023}&LISA~\cite{lai2024lisa}&74.9&79.1&72.3&65.1&70.8&58.1&67.9&70.6\\
        \rowcolor{cyan!15} RICE&Vicuna-7B~\cite{vicuna2023}&LISA~\cite{lai2024lisa}&76.3&80.3&75.1&67.4&72.7&60.6&69.0&73.4\\
        \multicolumn{3}{l}{\textcolor{ForestGreen}{\textbf{Avg: +2.00} ↑ Improvement over CLIP}}& \textcolor{ForestGreen}{+1.4} & \textcolor{ForestGreen}{+1.2} & \textcolor{ForestGreen}{+2.8} & \textcolor{ForestGreen}{+2.3} & \textcolor{ForestGreen}{+1.9} & \textcolor{ForestGreen}{+2.5} & \textcolor{ForestGreen}{+1.1} & \textcolor{ForestGreen}{+2.8} \\
        \midrule
        \multicolumn{11}{l}{\textit{\textbf{LLaVA-NeXT Framework}}} \\
        \midrule
        CLIP&Qwen2.5-7B&LISA~\cite{lai2024lisa}&81.8&84.0&79.1&76.6&80.5&70.9&77.3&78.5\\
        MLCD&Qwen2.5-7B&LISA~\cite{lai2024lisa}&82.8&84.6&80.2&77.4&81.6&73.1&78.5&79.7\\
        \rowcolor{cyan!15} RICE&Qwen2.5-7B&LISA~\cite{lai2024lisa}&\textbf{83.5}&\textbf{85.3}&\textbf{81.7}&\textbf{79.4}&\textbf{82.8}&\textbf{75.4}&\textbf{79.8}&\textbf{80.4}\\
        \multicolumn{3}{l}{ \textcolor{ForestGreen}{\textbf{Avg: +2.45} ↑ Improvement over CLIP }}  & \textcolor{ForestGreen}{+1.7} & \textcolor{ForestGreen}{+1.3} & \textcolor{ForestGreen}{+2.6} & \textcolor{ForestGreen}{+2.8} & \textcolor{ForestGreen}{+2.3} & \textcolor{ForestGreen}{+4.5} & \textcolor{ForestGreen}{+2.5} & \textcolor{ForestGreen}{+1.9} \\
        \multicolumn{3}{l}{ \textcolor{ForestGreen}{\textbf{Avg: +1.30} ↑ Improvement over MLCD }} & \textcolor{ForestGreen}{+0.7} & \textcolor{ForestGreen}{+0.7} & \textcolor{ForestGreen}{+1.5} & \textcolor{ForestGreen}{+2.0} & \textcolor{ForestGreen}{+1.2} & \textcolor{ForestGreen}{+2.3} & \textcolor{ForestGreen}{+1.3} & \textcolor{ForestGreen}{+0.7} \\
        \bottomrule
    \end{tabular}
}
\vspace{-2mm}
\caption{
Performance comparison of referring image segmentation across vision-language models. Results are reported as IoU scores (\%) on the refCOCO, refCOCO+, and refCOCOg benchmarks. Our RICE vision encoder consistently outperforms all competing approaches, achieving state-of-the-art results across all benchmarks when integrated with Qwen2.5-7B in the LLaVA-NeXT framework.
}
\label{tab:refcoco}
\vspace{-4mm}
\end{table*}

\begin{figure}[t!]
\centering
\includegraphics[width=0.99\linewidth]{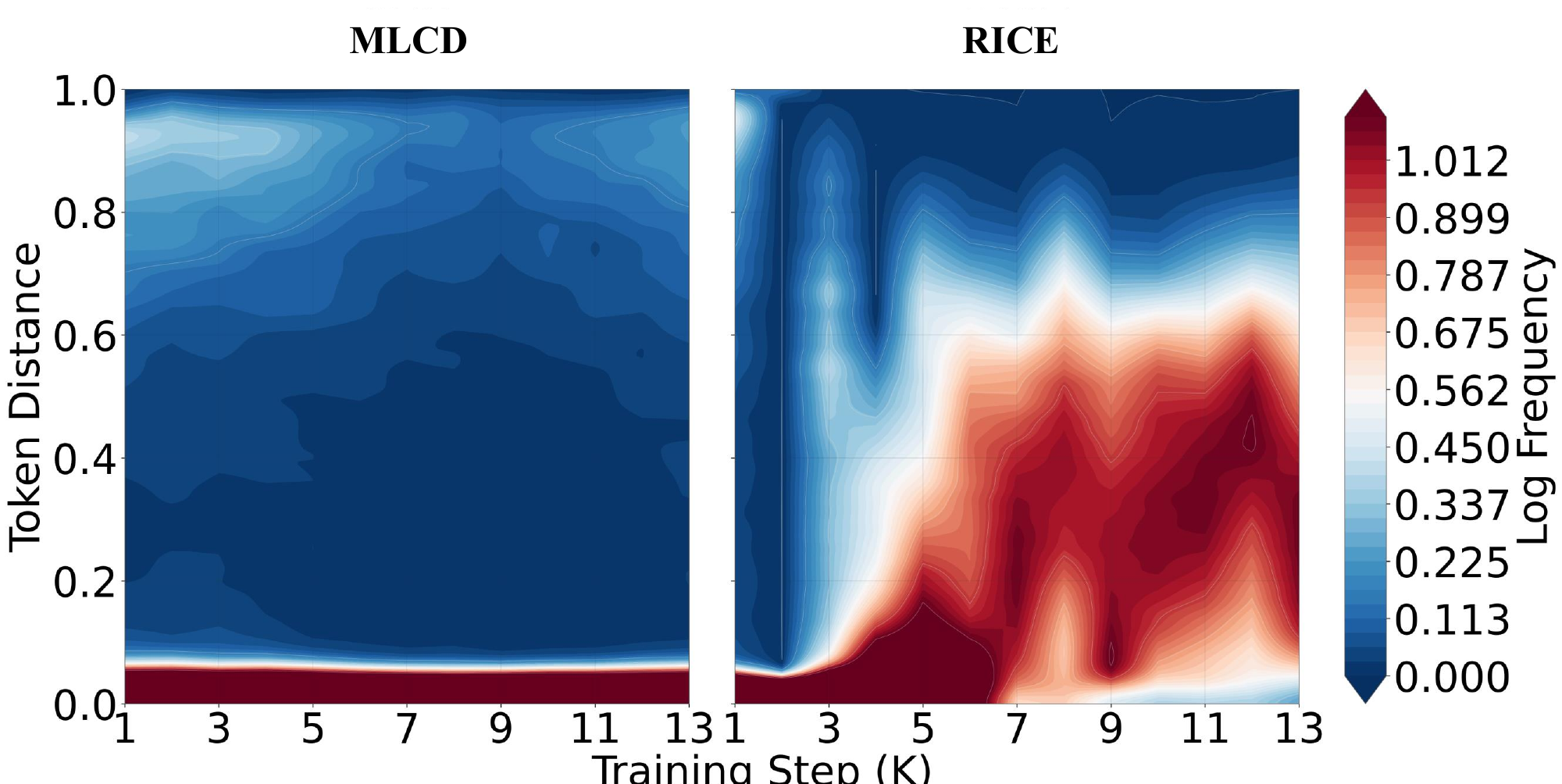}
\vspace{-2mm}
\caption{Token distance distributions observed during the training of MLCD and RICE models.}
\vspace{-4mm}
\label{fig:distance}
\end{figure}

\noindent{\bf Performance of Referring Segmentation.}
Table~\ref{tab:refcoco} presents an in-depth comparison of various vision-language model configurations using both the LLaVA-1.5 and LLaVA-NeXT (1.6) architectures. Within the LLaVA-1.5 framework, the RICE vision tower combined with Vicuna-7B consistently outperforms the standard CLIP vision encoder, achieving improvements of 1.4\%, 1.2\%, and 2.8\% on the refCOCO val, testA, and testB splits, respectively. Notably, in the more advanced LLaVA-NeXT framework with Qwen2.5-7B, RICE significantly surpasses the baseline MLCD method across all benchmarks. Specifically, RICE achieves relative gains of 0.7\% on refCOCO val, 0.7\% on refCOCO testA, and 1.5\% on refCOCO testB. This significant performance improvement is attributed to RICE's ability to more accurately express local semantic information. To further validate this point, we calculated pairwise distances between different image tokens based on the COCO dataset. As illustrated in Fig.~\ref{fig:distance}, RICE demonstrates superior capability to differentiate between visual tokens during training compared to MLCD, thereby enabling more precise object perception within the image.

\begin{table*}[!t]
    \centering
    \fontsize{8pt}{8pt}\selectfont
    \setlength\tabcolsep{4.3pt} 
    \renewcommand{\arraystretch}{1.3} 
    \begin{tabular}{lll|cc|cc|ccccccccc}
        \toprule
        \multicolumn{3}{c|}{\textbf{Configuration}} &
        \multicolumn{2}{c|}{\textbf{COCO}} &
        \multicolumn{2}{c|}{\textbf{LVIS}} &
        \multicolumn{8}{c}{\textbf{Roboflow100 Benchmarks}} \\
        \cmidrule(lr){1-3} \cmidrule(lr){4-5} \cmidrule(lr){6-7} \cmidrule(lr){8-15}
        \makecell[c]{Method} &
        \makecell[c]{Arch} &
        \makecell[c]{Res} &
        \makecell[c]{Det\\AP} &
        \makecell[c]{Seg\\AP} &
        \makecell[c]{Det\\AP} &
        \makecell[c]{Seg\\AP} &
        \makecell[c]{Aerial} &
        \makecell[c]{Video\\Games} &
        \makecell[c]{Micro-\\scopic} &
        \makecell[c]{Under\\Water} &
        \makecell[c]{Doc-\\uments} &
        \makecell[c]{Electro-\\magnetic} &
        \makecell[c]{Real\\world} &
        \makecell[c]{AVG} \\
        \midrule
        DINOv2 & ViT-B/14 & 518 & 31.6 & 24.3 & 18.7 & 14.1 & 2.3 & 14.3 & 10.6 & 19.9 & 18.8 & 15.3 & 26.8 & 15.4 \\
        SigLIP & ViT-B/16 & 512 & 35.0 & 28.1 & 21.8 & 17.3 & 9.4 & 29.5 & 20.0 & 29.4 & 23.4 & 18.6 & 38.0 & 24.1 \\
        MLCD & ViT-B/16 & 512 & 35.6 & 28.6 & 22.1 & 17.8 & 11.4 & 19.9 & 14.9 & 21.0 & 13.3 & 15.8 & 25.0 & 17.3 \\
        \rowcolor{cyan!15} RICE & ViT-B/16 & 512 & \textbf{38.9} & \textbf{31.5} & \textbf{26.5} & \textbf{21.4} & \textbf{14.9} & \textbf{31.7} & \textbf{23.4} & \textbf{30.7} & \textbf{27.0} & \textbf{18.7} & \textbf{39.1} & \textbf{26.5} \\
    \bottomrule
    \end{tabular}
\vspace{-2mm}
\caption{Performance comparison of vision encoders across multiple benchmarks, grouped by model size. Results are reported as Average Precision (\%) on COCO Detection, COCO Segmentation, LVIS Detection, LVIS Segmentation, and the Roboflow100-VL benchmark across various domains. Our proposed RICE vision encoder exhibits superior performance, consistently outperforming existing methods across these diverse evaluation settings.}
\label{tab:benchmark_results} 
\vspace{-4mm}
\end{table*}

\subsection{Detection Probe}
We evaluate RICE against several state-of-the-art pre-trained vision encoders across multiple benchmarks. All evaluations are conducted using the Cascade Mask R-CNN framework~\cite{he2017mask,cai2018cascade}  implemented in Detectron2~\cite{wu2019detectron2}. As summarized in Table~\ref{tab:benchmark_results}, experiments are performed on the COCO (80 classes) and LVIS (1203 classes) validation sets. Our feature pyramid is constructed directly from frozen backbone embeddings, employing max-pooling and unpooling operations to generate multi-scale feature maps at resolutions of $0.25\times$, $0.5\times$, $1\times$, $2\times$, and $4\times$.
RICE achieves superior performance across all evaluation metrics. On COCO, RICE attains 38.9\% detection AP and 31.5\% segmentation AP, surpassing the strongest baseline, SigLIP, by +3.9\% and +3.4\%, respectively. On LVIS, RICE achieves 26.5\% detection AP and 21.4\% segmentation AP, representing improvements of +4.7\% and +4.1\% over SigLIP. On the Roboflow100 benchmarks, RICE demonstrates exceptional generalization across diverse domains, achieving a 26.5\% average performance and notable gains in aerial imagery (+5.5\%) and microscopic analysis (+3.4\%). These consistent improvements on both natural images (COCO/LVIS) and specialized domains (Roboflow100) highlight RICE’s superior representational quality and its effectiveness for a wide range of computer vision applications.

\begin{table*}[!t]
    \centering
    \fontsize{8pt}{8pt}\selectfont
    \setlength\tabcolsep{4.3pt} 
    \renewcommand{\arraystretch}{1.3} 
    \begin{tabular}{l|ccc|ccc|ccc|ccc}
        \toprule
        \multirow{3}{*}{\textbf{Method}} &
        \multicolumn{3}{c|}{\textbf{LaSOT}} &
        \multicolumn{3}{c|}{\textbf{TrackingNet}} &
        \multicolumn{3}{c|}{\textbf{GOT-10k}} &
        \multicolumn{3}{c}{\textbf{TNL2k}} \\
        \cmidrule(lr){2-4} \cmidrule(lr){5-7} \cmidrule(lr){8-10} \cmidrule(lr){11-13}
        &
        \makecell[c]{Suc.} &
        \makecell[c]{Pre.} &
        \makecell[c]{Norm. Pre.} &
        \makecell[c]{Suc.} &
        \makecell[c]{Pre.} &
        \makecell[c]{Norm. Pre.} &
        \makecell[c]{AO} &
        \makecell[c]{SR-0.5} &
        \makecell[c]{SR-0.75} &
        \makecell[c]{Suc.} &
        \makecell[c]{Pre.} &
        \makecell[c]{Norm. Pre.} \\
        \midrule
        
        DINOv2 & 55.11 & 54.99 & 65.52 & 71.20 & 64.70 & 77.70 & 53.60 & 64.90 & 35.50 & 41.95 & 36.03 & 57.40 \\
        SigLIP & 55.52 & 56.16 & 65.33 & 72.60 & 66.70 & 78.70 & 53.50 & 63.10 & 35.40 & 43.90 & 39.06 & 59.03 \\
        MLCD & 58.05 & 60.75 & 68.31 & 75.30 & 70.20 & 80.80 & 53.80 & 62.80 & 39.70 & 45.22 & 40.62 & 60.64 \\
        
        \rowcolor{cyan!15} RICE & \textbf{60.24} & \textbf{63.16} & \textbf{69.66} & \textbf{76.30} & \textbf{71.80} & \textbf{81.30} & \textbf{55.40} & \textbf{63.50} & \textbf{41.60} & \textbf{45.70} & \textbf{41.95} & \textbf{61.18} \\
        
        \bottomrule
    \end{tabular}
    \vspace{-2mm}
    \caption{Performance comparison of different vision encoders on video tracking benchmarks. Results are reported on LaSOT, TrackingNet, GOT-10k, and TNL2k datasets. Our proposed RICE vision encoder demonstrates superior performance across all metrics.}
    \vspace{-2mm}
    \label{tab:tracking_results_method_double} 
\end{table*}

\begin{figure*}[t!]
\centering
\includegraphics[width=0.98\linewidth]{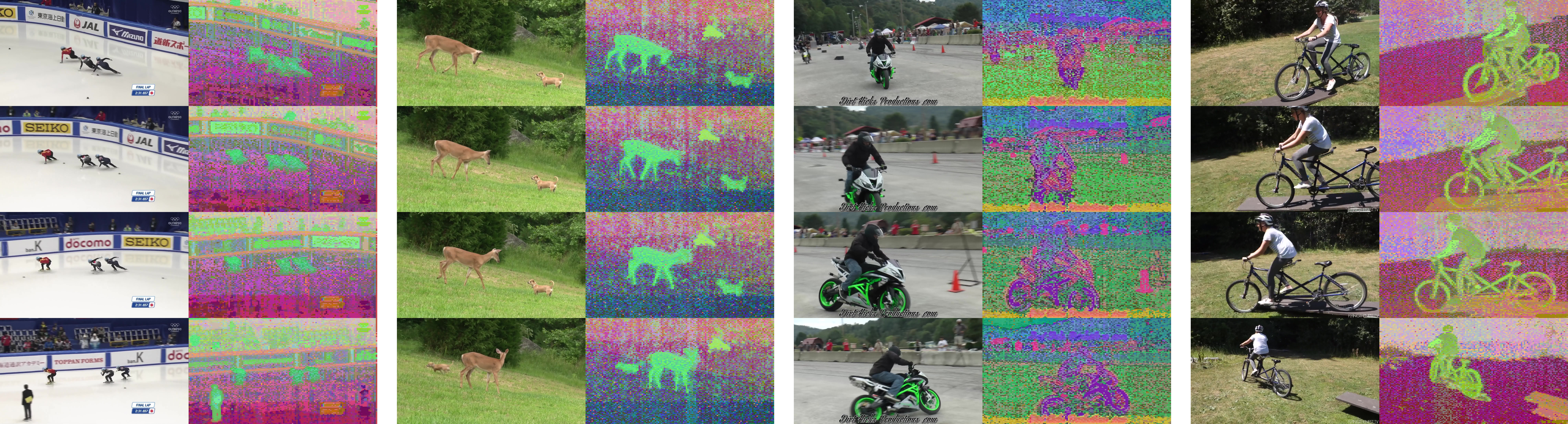}
\vspace{-2mm}
\caption{Tracking PCA. Using 2048-resolution images as input to a ViT-B/16 model, we project token features onto RGB channels via PCA to visualize the semantic structure. Sequential frames (arranged vertically) illustrate the evolution of model attention, consistently highlighting salient objects across time. The visualization reveals stable color patterns for tracked entities such as ice skaters, deers, motorcyclists, and cyclists, demonstrating the model’s ability to maintain semantic focus throughout the sequence.}
\label{fig:tracking}
\vspace{-2mm}
\end{figure*}

\newcolumntype{C}[1]{>{\centering\arraybackslash}p{#1}}

\begin{table*}[!t]
    \centering
    \newcommand{\mycolwidth}{1em} 
    \begin{subtable}[t]{0.24\textwidth}
        \centering
\fontsize{7pt}{7.5pt}\selectfont    
\setlength\tabcolsep{5.2pt}           
\renewcommand{\arraystretch}{1.2}    
        \begin{tabular}{l|*{4}{C{\mycolwidth}}}
            \toprule
            Testset & 1 & 5 & 10 & 20 \\
            \midrule
            COCO-Det & 31.4 & 34.5 & \textbf{35.2} & 34.7 \\
            COCO-Seg & 20.7 & 27.1 & \textbf{28.6} & 27.2 \\
            LVIS-Det & 17.5 & 21.2 & \textbf{23.1} & 22.4 \\
            LVIS-Seg & 14.3 & 17.3 & \textbf{19.9} & 19.0 \\
            \bottomrule
        \end{tabular}
        \captionsetup{font=footnotesize}
        \caption{Sampled boxes ($N$).}
        \label{tab:ablation_bboxes_r1}
    \end{subtable}%
    \hfill
    \begin{subtable}[t]{0.24\textwidth}
        \centering
        \fontsize{7pt}{7.5pt}\selectfont
        \setlength\tabcolsep{5.2pt}
        \renewcommand{\arraystretch}{1.2}
        \begin{tabular}{l|*{4}{C{\mycolwidth}}}
            \toprule
            Testset  & 200K & 500K & 1M & 2M \\
            \midrule
            COCO-Det & 32.8 & 33.6 & \textbf{35.2} & 34.9 \\
            COCO-Seg & 24.3 & 26.8 & \textbf{28.6} & 27.9 \\
            LVIS-Det & 18.9 & 21.6 & 23.1 & \textbf{24.0} \\
            LVIS-Seg & 15.4 & 17.8 & \textbf{19.9} & 18.7 \\
            \bottomrule
        \end{tabular}
        \captionsetup{font=footnotesize}
        \caption{Cluster centers ($K$).}
        \label{tab:ablation_clusters_r1}
    \end{subtable}%
    \hfill
    \begin{subtable}[t]{0.24\textwidth}
        \centering
        \fontsize{7pt}{7.5pt}\selectfont
        \setlength\tabcolsep{5pt}
        \renewcommand{\arraystretch}{1.2}
        \begin{tabular}{l|*{4}{C{\mycolwidth}}}
            \toprule
            Testset  & 0.05 & 0.1 & 0.2 & 0.5 \\
            \midrule
            COCO-Det & \textbf{38.8} & 35.2 & 35.1 & 33.7 \\
            COCO-Seg & 27.3 & \textbf{28.6} & 28.4 & 27.4 \\
            LVIS-Det & 22.1 & \textbf{23.1} & 22.4 & 21.6 \\
            LVIS-Seg & 19.8 & \textbf{19.9} & 19.1 & 18.6 \\
            \bottomrule
        \end{tabular}
        \captionsetup{font=footnotesize}
        \caption{Negative sampling rate ($\rho$).}
        \label{tab:ablation_sampling_r1}
    \end{subtable}%
    \hfill
    \begin{subtable}[t]{0.24\textwidth}
        \centering
        \fontsize{7pt}{7.5pt}\selectfont
        \setlength\tabcolsep{5.2pt}
        \renewcommand{\arraystretch}{1.2}
        \begin{tabular}{l|*{4}{C{\mycolwidth}}}
            \toprule
            Testset& 3 & 5 & 10 & 20 \\
            \midrule
            DocVQA & 53.1 & 54.3 & \textbf{54.4} & 53.0 \\
            ChartQA & 62.7 & 62.2 & \textbf{63.0} & 62.9 \\
            InfoVQA & 37.3 & 37.5 & 38.2 & \textbf{38.4} \\
            OCRBench & 391 & 392 & \textbf{403} & 401 \\
            \bottomrule
        \end{tabular}
        \captionsetup{font=footnotesize}
        \caption{Positive OCR labels ($M$).}
        \label{tab:ablation_ocr_labels_r2}
    \end{subtable}
\vspace{-2mm}
\caption{Ablation studies on key hyperparameters using the ViT-B/16 backbone and 10\% of the pretraining dataset to investigate optimal settings for detection, segmentation, and OCR tasks.
}
\vspace{-3mm}
\label{tab:ablation_studies}
\end{table*}

\subsection{Tracking Probe}
We evaluate the temporal matching capability of local features within the general video object tracking framework OSTrack~\cite{OSTrack}, adopting an attention probing approach to compare the four pre-trained models. Specifically, we insert two standard vision transformer blocks between the frozen backbone and the prediction head to enhance information exchange between the template and search images, while only passing the search image features to the position prediction network. For GOT-10k~\cite{got10k}, models are trained for 100 epochs on the GOT-10k training set. For evaluations on LaSOT~\cite{lasot}, TrackingNet~\cite{tknet}, and TNL2K~\cite{wang2021towards}, we train for 300 epochs using the combined training sets from COCO, TrackingNet, GOT-10k, and LaSOT. As shown in Tab.~\ref{tab:tracking_results_method_double}, RICE achieves the best performance across all metrics on the above four datasets.

To further interpret the model’s behavior, we visualize semantic features by projecting token representations from the ViT-B/16 model onto RGB channels via PCA. The coherence of color patterns across frames demonstrates the model’s ability to dynamically maintain focus on salient objects while adapting to movement and contextual changes. In Fig.~\ref{fig:tracking}, we visualize frames from diverse videos, where each vertical sequence depicts consecutive frames, the original images on the left and their corresponding PCA visualizations on the right. RICE keeps stable attention on tracked entities such as ice skaters, deers, motorcyclists, and cyclists, not only confirming the robustness of the model’s object tracking capabilities but also illustrating the value of PCA-based visualization for revealing semantic feature distributions and temporal attention dynamics.

\subsection{Ablation Study}
We conducted extensive ablation studies on key hyperparameters using the ViT-B/16 backbone and 10\% of the pretraining dataset, as presented in Tab.~\ref{tab:ablation_bboxes_r1}--\ref{tab:ablation_ocr_labels_r2}. Specifically, Tab.~\ref{tab:ablation_bboxes_r1} demonstrates that utilizing 10 object bounding boxes yields the best performance across all detection and segmentation benchmarks. For cluster centers (Tab.\ref{tab:ablation_clusters_r1}), we observe optimal performance with 1--2 million centers, where COCO benchmarks peak at 1M and LVIS-Det achieves its highest results at 2M. As shown in Tab.\ref{tab:ablation_sampling_r1}, lower negative sampling rates lead to superior outcomes, with COCO-Det attaining the best performance at a rate of 0.05, while other metrics perform best at 0.1. Our experimental framework also includes analyses on the impact of positive OCR labels, as reported in Tab.~\ref{tab:ablation_ocr_labels_r2}.  

\section{Conclusions}
In this paper, we propose a novel Region-Aware Cluster Discrimination approach to advance region-aware object recognition and OCR capabilities. We first construct a billion-scale candidate region dataset and introduce the RICE model, which incorporates a Region Transformer layer to capture regional semantic information. Furthermore, we develop an efficient region-based classification loss that jointly supports object and OCR learning, enabling scalable and distributed training on large-scale data. Extensive experiments demonstrate that RICE consistently outperforms prior methods across a range of downstream tasks, including segmentation, dense detection, and visual perception for MLLMs.

\clearpage

{
\small
\bibliographystyle{ieeenat_fullname}
\bibliography{main}
}

\end{document}


%% file: main.bbl
\begin{thebibliography}{85}
\providecommand{\natexlab}[1]{#1}
\providecommand{\url}[1]{\texttt{#1}}
\expandafter\ifx\csname urlstyle\endcsname\relax
  \providecommand{\doi}[1]{doi: #1}\else
  \providecommand{\doi}{doi: \begingroup \urlstyle{rm}\Url}\fi

\bibitem[Alayrac et~al.(2022)Alayrac, Donahue, Luc, Miech, Barr, Hasson, Lenc, Mensch, Millican, Reynolds, et~al.]{alayrac2022flamingo}
Jean-Baptiste Alayrac, Jeff Donahue, Pauline Luc, Antoine Miech, Iain Barr, Yana Hasson, Karel Lenc, Arthur Mensch, Katherine Millican, Malcolm Reynolds, et~al.
\newblock {Flamingo: a Visual Language Model for Few-Shot Learning}.
\newblock In \emph{NeurIPS}, 2022.

\bibitem[An et~al.(2022)An, Deng, Guo, Feng, Zhu, Yang, and Liu]{An2022PartialFC}
Xiang An, Jiankang Deng, Jia Guo, Ziyong Feng, XuHan Zhu, Jing Yang, and Tongliang Liu.
\newblock {Killing Two Birds with One Stone: Efficient and Robust Training of Face Recognition CNNs by Partial FC}.
\newblock In \emph{{CVPR}}, 2022.

\bibitem[An et~al.(2023)An, Deng, Yang, Li, Feng, Guo, Yang, and Liu]{an2023unicom}
Xiang An, Jiankang Deng, Kaicheng Yang, Jiawei Li, Ziyong Feng, Jia Guo, Jing Yang, and Tongliang Liu.
\newblock {Unicom: Universal and Compact Representation Learning for Image Retrieval}.
\newblock In \emph{{ICLR}}, 2023.

\bibitem[An et~al.(2024)An, Yang, Dai, Feng, and Deng]{anxiang_2024_mlcd}
Xiang An, Kaicheng Yang, Xiangzi Dai, Ziyong Feng, and Jiankang Deng.
\newblock {Multi-label Cluster Discrimination for Visual Representation Learning}.
\newblock In \emph{{ECCV}}, 2024.

\bibitem[Asano et~al.(2020)Asano, Rupprecht, and Vedaldi]{asano2019self}
Yuki~Markus Asano, Christian Rupprecht, and Andrea Vedaldi.
\newblock {Self-Labelling via Simultaneous Clustering and Representation Learning}.
\newblock In \emph{{ICLR}}, 2020.

\bibitem[Bai et~al.(2023{\natexlab{a}})Bai, Bai, Chu, Cui, Dang, Deng, Fan, Ge, Han, Huang, et~al.]{qwen1}
Jinze Bai, Shuai Bai, Yunfei Chu, Zeyu Cui, Kai Dang, Xiaodong Deng, Yang Fan, Wenbin Ge, Yu Han, Fei Huang, et~al.
\newblock {Qwen technical report}.
\newblock 2023{\natexlab{a}}.

\bibitem[Bai et~al.(2023{\natexlab{b}})Bai, Bai, Yang, Wang, Tan, Wang, Lin, Zhou, and Zhou]{Qwen-VL}
Jinze Bai, Shuai Bai, Shusheng Yang, Shijie Wang, Sinan Tan, Peng Wang, Junyang Lin, Chang Zhou, and Jingren Zhou.
\newblock {Qwen-VL: A Versatile Vision-Language Model for Understanding, Localization, Text Reading, and Beyond}.
\newblock \emph{arXiv:2308.12966}, 2023{\natexlab{b}}.

\bibitem[Byeon et~al.(2022)Byeon, Park, Kim, Lee, Baek, and Kim]{kakaobrain2022coyo-700m}
Minwoo Byeon, Beomhee Park, Haecheon Kim, Sungjun Lee, Woonhyuk Baek, and Saehoon Kim.
\newblock {COYO-700M: Image-Text Pair Dataset}, 2022.

\bibitem[Cai and Vasconcelos(2018)]{cai2018cascade}
Zhaowei Cai and Nuno Vasconcelos.
\newblock {Cascade R-CNN: Delving Into High Quality Object Detection}.
\newblock In \emph{{CVPR}}, 2018.

\bibitem[Caron et~al.(2018)Caron, Bojanowski, Joulin, and Douze]{caron2018deep}
Mathilde Caron, Piotr Bojanowski, Armand Joulin, and Matthijs Douze.
\newblock {Deep Clustering for Unsupervised Learning of Visual Features}.
\newblock In \emph{{ECCV}}, 2018.

\bibitem[Caron et~al.(2020)Caron, Misra, Mairal, Goyal, Bojanowski, and Joulin]{caron2020unsupervised}
Mathilde Caron, Ishan Misra, Julien Mairal, Priya Goyal, Piotr Bojanowski, and Armand Joulin.
\newblock {Unsupervised Learning of Visual Features by Contrasting Cluster Assignments}.
\newblock In \emph{{NeurIPS}}, 2020.

\bibitem[Chen et~al.(2024{\natexlab{a}})Chen, Yu, Shen, Yuille, and Chen]{chen2024vitamin}
Jieneng Chen, Qihang Yu, Xiaohui Shen, Alan Yuille, and Liang-Chieh Chen.
\newblock {ViTamin: Designing Scalable Vision Models in the Vision-language Era}.
\newblock In \emph{{CVPR}}, 2024{\natexlab{a}}.

\bibitem[Chen et~al.(2024{\natexlab{b}})Chen, Li, Dong, Zhang, Zang, Chen, Duan, Wang, Qiao, Lin, and Zhao]{vlm_mmstar}
Lin Chen, Jinsong Li, Xiaoyi Dong, Pan Zhang, Yuhang Zang, Zehui Chen, Haodong Duan, Jiaqi Wang, Yu Qiao, Dahua Lin, and Feng Zhao.
\newblock {Are We on the Right Way for Evaluating Large Vision-Language Models?}
\newblock In \emph{{NeurIPS}}, 2024{\natexlab{b}}.

\bibitem[Chen et~al.(2024{\natexlab{c}})Chen, Wang, Cao, Liu, Gao, Cui, Zhu, Ye, Tian, Liu, et~al.]{chen2024expanding}
Zhe Chen, Weiyun Wang, Yue Cao, Yangzhou Liu, Zhangwei Gao, Erfei Cui, Jinguo Zhu, Shenglong Ye, Hao Tian, Zhaoyang Liu, et~al.
\newblock {Expanding Performance Boundaries of Open-Source Multimodal Models with Model, Data, and Test-Time Scaling}.
\newblock \emph{arXiv:2412.05271}, 2024{\natexlab{c}}.

\bibitem[Chiang et~al.(2023)Chiang, Li, Lin, Sheng, Wu, Zhang, Zheng, Zhuang, Zhuang, Gonzalez, Stoica, and Xing]{vicuna2023}
Wei-Lin Chiang, Zhuohan Li, Zi Lin, Ying Sheng, Zhanghao Wu, Hao Zhang, Lianmin Zheng, Siyuan Zhuang, Yonghao Zhuang, Joseph~E. Gonzalez, Ion Stoica, and Eric~P. Xing.
\newblock {Vicuna: An Open-Source Chatbot Impressing GPT-4 with 90\%* ChatGPT Quality}.
\newblock 2023.

\bibitem[Deng et~al.(2019)Deng, Guo, Xue, and Zafeiriou]{deng2019arcface}
Jiankang Deng, Jia Guo, Niannan Xue, and Stefanos Zafeiriou.
\newblock {Arcface: Additive angular margin loss for deep face recognition}.
\newblock In \emph{{CVPR}}, 2019.

\bibitem[Douze et~al.(2025)Douze, Guzhva, Deng, Johnson, Szilvasy, Mazare, Lomeli, Hosseini, and Jegou]{douze2025faisslibrary}
Matthijs Douze, Alexandr Guzhva, Chengqi Deng, Jeff Johnson, Gergely Szilvasy, Pierre-Emmanuel Mazare, Maria Lomeli, Lucas Hosseini, and Herve Jegou.
\newblock {The Faiss library}.
\newblock \emph{arXiv:2401.08281}, 2025.

\bibitem[Du et~al.(2020)Du, Li, Guo, Yin, Liu, Zhou, Bai, Yu, Yang, Dang, et~al.]{du2020pp}
Yuning Du, Chenxia Li, Ruoyu Guo, Xiaoting Yin, Weiwei Liu, Jun Zhou, Yifan Bai, Zilin Yu, Yehua Yang, Qingqing Dang, et~al.
\newblock {PP-OCR: A Practical Ultra Lightweight OCR System}.
\newblock \emph{arXiv:2009.09941}, 2020.

\bibitem[El-Nouby et~al.(2024)El-Nouby, Klein, Zhai, Bautista, Shankar, Toshev, Susskind, and Joulin]{aim}
Alaaeldin El-Nouby, Michal Klein, Shuangfei Zhai, Miguel~\'{A}ngel Bautista, Vaishaal Shankar, Alexander~T Toshev, Joshua~M. Susskind, and Armand Joulin.
\newblock {Scalable Pre-training of Large Autoregressive Image Models}.
\newblock \emph{PMLR}, 2024.

\bibitem[Fan et~al.(2019)Fan, Lin, Yang, Chu, Deng, Yu, Bai, Xu, Liao, and Ling]{lasot}
Heng Fan, Liting Lin, Fan Yang, Peng Chu, Ge Deng, Sijia Yu, Hexin Bai, Yong Xu, Chunyuan Liao, and Haibin Ling.
\newblock Lasot: A high-quality benchmark for large-scale single object tracking.
\newblock In \emph{CVPR}, 2019.

\bibitem[Fang et~al.(2024)Fang, Jose, Jain, Schmidt, Toshev, and Shankar]{DFN5B}
Alex Fang, Albin~Madappally Jose, Amit Jain, Ludwig Schmidt, Alexander~T Toshev, and Vaishaal Shankar.
\newblock {Data Filtering Networks}.
\newblock In \emph{{ICLR}}, 2024.

\bibitem[Fini et~al.(2025)Fini, Shukor, Li, Dufter, Klein, Haldimann, Aitharaju, da~Costa, B{\'e}thune, Gan, Toshev, Eichner, Nabi, Yang, Susskind, and El-Nouby]{fini2025multimodal}
Enrico Fini, Mustafa Shukor, Xiujun Li, Philipp Dufter, Michal Klein, David Haldimann, Sai Aitharaju, Victor G~Turrisi da Costa, Louis B{\'e}thune, Zhe Gan, Alexander Toshev, Marcin Eichner, Moin Nabi, Yinfei Yang, Joshua Susskind, and Alaaeldin El-Nouby.
\newblock Multimodal autoregressive pre-training of large vision encoders.
\newblock In \emph{CVPR}, 2025.

\bibitem[Gu et~al.(2024)Gu, Yang, An, Feng, Liu, Cai, and Deng]{gu2024rwkvclip}
Tiancheng Gu, Kaicheng Yang, Xiang An, Ziyong Feng, Dongnan Liu, Weidong Cai, and Jiankang Deng.
\newblock Rwkv-clip: A robust vision-language representation learner.
\newblock In \emph{EMNLP}, 2024.

\bibitem[He et~al.(2017)He, Gkioxari, Doll{\'a}r, and Girshick]{he2017mask}
Kaiming He, Georgia Gkioxari, Piotr Doll{\'a}r, and Ross Girshick.
\newblock {Mask R-CNN}.
\newblock In \emph{{ICCV}}, 2017.

\bibitem[Huang et~al.(2019)Huang, Zhao, and Huang]{got10k}
Lianghua Huang, Xin Zhao, and Kaiqi Huang.
\newblock {GOT-10k: A Large High-Diversity Benchmark for Generic Object Tracking in the Wild}.
\newblock \emph{TPAMI}, 2019.

\bibitem[Hui et~al.(2024)Hui, Yang, Cui, Yang, Liu, Zhang, Liu, Zhang, Yu, Dang, et~al.]{qwen2.5}
Binyuan Hui, Jian Yang, Zeyu Cui, Jiaxi Yang, Dayiheng Liu, Lei Zhang, Tianyu Liu, Jiajun Zhang, Bowen Yu, Kai Dang, et~al.
\newblock {Qwen2.5-Coder Technical Report}.
\newblock \emph{arXiv:2409.12186}, 2024.

\bibitem[Ilharco et~al.(2021)Ilharco, Wortsman, Wightman, Gordon, Carlini, Taori, Dave, Shankar, Namkoong, Miller, Hajishirzi, Farhadi, and Schmidt]{openclip}
Gabriel Ilharco, Mitchell Wortsman, Ross Wightman, Cade Gordon, Nicholas Carlini, Rohan Taori, Achal Dave, Vaishaal Shankar, Hongseok Namkoong, John Miller, Hannaneh Hajishirzi, Ali Farhadi, and Ludwig Schmidt.
\newblock {OpenCLIP}.
\newblock 2021.

\bibitem[Kazemzadeh et~al.(2014)Kazemzadeh, Ordonez, Matten, and Berg]{kazemzadeh2014referitgame}
Sahar Kazemzadeh, Vicente Ordonez, Mark Matten, and Tamara Berg.
\newblock {ReferItGame: Referring to Objects in Photographs of Natural Scenes}.
\newblock In \emph{{EMNLP}}, 2014.

\bibitem[Kembhavi et~al.(2016)Kembhavi, Salvato, Kolve, Seo, Hajishirzi, and Farhadi]{vlm_ai2d}
Aniruddha Kembhavi, Mike Salvato, Eric Kolve, Minjoon Seo, Hannaneh Hajishirzi, and Ali Farhadi.
\newblock {A Diagram Is Worth A Dozen Images}.
\newblock In \emph{{ECCV}}, 2016.

\bibitem[Kirillov et~al.(2023)Kirillov, Mintun, Ravi, Mao, Rolland, Gustafson, Xiao, Whitehead, Berg, Lo, Dollar, and Girshick]{sam}
Alexander Kirillov, Eric Mintun, Nikhila Ravi, Hanzi Mao, Chloe Rolland, Laura Gustafson, Tete Xiao, Spencer Whitehead, Alexander~C. Berg, Wan-Yen Lo, Piotr Dollar, and Ross Girshick.
\newblock {Segment Anything}.
\newblock In \emph{{ICCV}}, 2023.

\bibitem[Lai et~al.(2024)Lai, Tian, Chen, Li, Yuan, Liu, and Jia]{lai2024lisa}
Xin Lai, Zhuotao Tian, Yukang Chen, Yanwei Li, Yuhui Yuan, Shu Liu, and Jiaya Jia.
\newblock {Lisa: Reasoning segmentation via large language model}.
\newblock In \emph{{CVPR}}, 2024.

\bibitem[Li et~al.(2024)Li, Zhang, Guo, Zhang, Li, Zhang, Zhang, Zhang, Li, Liu, et~al.]{li2024llavaov}
Bo Li, Yuanhan Zhang, Dong Guo, Renrui Zhang, Feng Li, Hao Zhang, Kaichen Zhang, Peiyuan Zhang, Yanwei Li, Ziwei Liu, et~al.
\newblock {LLaVA-OneVision: Easy Visual Task Transfer}.
\newblock \emph{arXiv:2408.03326}, 2024.

\bibitem[Li et~al.(2025)Li, Zhang, Zhang, Zhang, Li, Li, MA, and Li]{li2024llavanext}
Feng Li, Renrui Zhang, Hao Zhang, Yuanhan Zhang, Bo Li, Wei Li, Zejun MA, and Chunyuan Li.
\newblock {LLaVA-NeXT-Interleave: Tackling Multi-image, Video, and 3D in Large Multimodal Models}.
\newblock In \emph{{ICLR}}, 2025.

\bibitem[Li et~al.(2022{\natexlab{a}})Li, Li, Xiong, and Hoi]{li2022blip}
Junnan Li, Dongxu Li, Caiming Xiong, and Steven Hoi.
\newblock {BLIP: Bootstrapping Language-Image Pre-training for Unified Vision-Language Understanding and Generation}.
\newblock In \emph{{ICML}}, 2022{\natexlab{a}}.

\bibitem[Li et~al.(2023{\natexlab{a}})Li, Li, Savarese, and Hoi]{li2023blip}
Junnan Li, Dongxu Li, Silvio Savarese, and Steven Hoi.
\newblock {BLIP-2: Bootstrapping Language-Image Pre-training with Frozen Image Encoders and Large Language Models}.
\newblock In \emph{{ICML}}, 2023{\natexlab{a}}.

\bibitem[Li et~al.(2022{\natexlab{b}})Li, Zhang, Zhang, Yang, Li, Zhong, Wang, Yuan, Zhang, Hwang, Chang, and Gao]{li2021grounded}
Liunian~Harold Li, Pengchuan Zhang, Haotian Zhang, Jianwei Yang, Chunyuan Li, Yiwu Zhong, Lijuan Wang, Lu Yuan, Lei Zhang, Jenq-Neng Hwang, Kai-Wei Chang, and Jianfeng Gao.
\newblock {Grounded Language-Image Pre-training}.
\newblock In \emph{{CVPR}}, 2022{\natexlab{b}}.

\bibitem[Li et~al.(2023{\natexlab{b}})Li, Du, Zhou, Wang, Zhao, and Wen]{vlm_pope}
Yifan Li, Yifan Du, Kun Zhou, Jinpeng Wang, Wayne~Xin Zhao, and Ji-Rong Wen.
\newblock {Evaluating Object Hallucination in Large Vision-Language Models}.
\newblock In \emph{{EMNLP}}, 2023{\natexlab{b}}.

\bibitem[Liu et~al.(2024{\natexlab{a}})Liu, Li, Li, and Lee]{liu2024llava_V1_5}
Haotian Liu, Chunyuan Li, Yuheng Li, and Yong~Jae Lee.
\newblock {Improved Baselines with Visual Instruction Tuning}.
\newblock In \emph{{CVPR}}, 2024{\natexlab{a}}.

\bibitem[Liu et~al.(2024{\natexlab{b}})Liu, Li, Li, Li, Zhang, Shen, and Lee]{liu2024llavanext}
Haotian Liu, Chunyuan Li, Yuheng Li, Bo Li, Yuanhan Zhang, Sheng Shen, and Yong~Jae Lee.
\newblock {LLaVA-NeXT: Improved reasoning, OCR, and world knowledge}, 2024{\natexlab{b}}.

\bibitem[Liu et~al.(2023)Liu, Duan, Zhang, Li, Zhang, Zhao, Yuan, Wang, He, Liu, et~al.]{vlm_mmbench}
Yuan Liu, Haodong Duan, Yuanhan Zhang, Bo Li, Songyang Zhang, Wangbo Zhao, Yike Yuan, Jiaqi Wang, Conghui He, Ziwei Liu, et~al.
\newblock {MMBench: Is Your Multi-modal Model an All-around Player?}
\newblock \emph{arXiv:2307.06281}, 2023.

\bibitem[Liu et~al.(2024{\natexlab{c}})Liu, Li, Huang, Yang, Yu, Li, Yin, Liu, Jin, and Bai]{vlm_ocrbench}
Yuliang Liu, Zhang Li, Mingxin Huang, Biao Yang, Wenwen Yu, Chunyuan Li, Xu-Cheng Yin, Cheng-Lin Liu, Lianwen Jin, and Xiang Bai.
\newblock {OCRBench: On the Hidden Mystery of OCR in Large Multimodal Models}.
\newblock \emph{Science China Information Sciences}, 2024{\natexlab{c}}.

\bibitem[Liu et~al.(2022)Liu, Mao, Wu, Feichtenhofer, Darrell, and Xie]{convnext}
Zhuang Liu, Hanzi Mao, Chao-Yuan Wu, Christoph Feichtenhofer, Trevor Darrell, and Saining Xie.
\newblock {A ConvNet for the 2020s}.
\newblock In \emph{{CVPR}}, 2022.

\bibitem[Loshchilov and Hutter(2018)]{loshchilov2018decoupled}
Ilya Loshchilov and Frank Hutter.
\newblock {Decoupled Weight Decay Regularization}.
\newblock In \emph{{ICLR}}, 2018.

\bibitem[Masry et~al.(2022)Masry, Do, Tan, Joty, and Hoque]{vlm_chartqa}
Ahmed Masry, Xuan~Long Do, Jia~Qing Tan, Shafiq Joty, and Enamul Hoque.
\newblock {ChartQA: A Benchmark for Question Answering about Charts with Visual and Logical Reasoning}.
\newblock In \emph{{ACL Findings}}, 2022.

\bibitem[Mathew et~al.(2021)Mathew, Karatzas, and Jawahar]{vlm_docvqa}
Minesh Mathew, Dimosthenis Karatzas, and CV Jawahar.
\newblock {DocVQA: A Dataset for VQA on Document Images}.
\newblock In \emph{{WACV}}, 2021.

\bibitem[Mathew et~al.(2022)Mathew, Bagal, Tito, Karatzas, Valveny, and Jawahar]{vlm_infovqa}
Minesh Mathew, Viraj Bagal, Rub{\`e}n Tito, Dimosthenis Karatzas, Ernest Valveny, and CV Jawahar.
\newblock {Infographicvqa}.
\newblock In \emph{{WACV}}, 2022.

\bibitem[Muller et~al.(2018)Muller, Bibi, Giancola, Alsubaihi, and Ghanem]{tknet}
Matthias Muller, Adel Bibi, Silvio Giancola, Salman Alsubaihi, and Bernard Ghanem.
\newblock Trackingnet: A large-scale dataset and benchmark for object tracking in the wild.
\newblock In \emph{ECCV}, 2018.

\bibitem[Oquab et~al.(2024)Oquab, Darcet, Moutakanni, Vo, Szafraniec, Khalidov, Fernandez, Haziza, Massa, El-Nouby, Assran, Ballas, Galuba, Howes, Huang, Li, Misra, Rabbat, Sharma, Synnaeve, Xu, Jegou, Mairal, Labatut, Joulin, and Bojanowski]{oquab2024dinov2learningrobustvisual}
Maxime Oquab, Timothée Darcet, Théo Moutakanni, Huy Vo, Marc Szafraniec, Vasil Khalidov, Pierre Fernandez, Daniel Haziza, Francisco Massa, Alaaeldin El-Nouby, Mahmoud Assran, Nicolas Ballas, Wojciech Galuba, Russell Howes, Po-Yao Huang, Shang-Wen Li, Ishan Misra, Michael Rabbat, Vasu Sharma, Gabriel Synnaeve, Hu Xu, Hervé Jegou, Julien Mairal, Patrick Labatut, Armand Joulin, and Piotr Bojanowski.
\newblock {DINOv2: Learning Robust Visual Features without Supervision}.
\newblock \emph{PMLR}, 2024.

\bibitem[Radenovic et~al.(2023)Radenovic, Dubey, Kadian, Mihaylov, Vandenhende, Patel, Wen, Ramanathan, and Mahajan]{radenovic2023filtering}
Filip Radenovic, Abhimanyu Dubey, Abhishek Kadian, Todor Mihaylov, Simon Vandenhende, Yash Patel, Yi Wen, Vignesh Ramanathan, and Dhruv Mahajan.
\newblock {Filtering, Distillation, and Hard Negatives for Vision-Language Pre-Training}.
\newblock In \emph{{CVPR}}, 2023.

\bibitem[Radford et~al.(2021)Radford, Kim, Hallacy, Ramesh, Goh, Agarwal, Sastry, Askell, Mishkin, Clark, Krueger, and Sutskever]{clip_icml}
Alec Radford, Jong~Wook Kim, Chris Hallacy, Aditya Ramesh, Gabriel Goh, Sandhini Agarwal, Girish Sastry, Amanda Askell, Pamela Mishkin, Jack Clark, Gretchen Krueger, and Ilya Sutskever.
\newblock {Learning Transferable Visual Models From Natural Language Supervision}.
\newblock In \emph{{ICML}}, 2021.

\bibitem[Ramanathan et~al.(2021)Ramanathan, Wang, and Mahajan]{ramanathan2021predet}
Vignesh Ramanathan, Rui Wang, and Dhruv Mahajan.
\newblock {PreDet: Large-Scale Weakly Supervised Pre-Training for Detection}.
\newblock In \emph{{ICCV}}, 2021.

\bibitem[Rasheed et~al.(2024)Rasheed, Maaz, Shaji, Shaker, Khan, Cholakkal, Anwer, Xing, Yang, and Khan]{rasheed2024glamm}
Hanoona Rasheed, Muhammad Maaz, Sahal Shaji, Abdelrahman Shaker, Salman Khan, Hisham Cholakkal, Rao~M Anwer, Eric Xing, Ming-Hsuan Yang, and Fahad~S Khan.
\newblock {GLaMM: Pixel Grounding Large Multimodal Model}.
\newblock In \emph{{CVPR}}, 2024.

\bibitem[Redmon et~al.(2016)Redmon, Divvala, Girshick, and Farhadi]{redmon2016you}
Joseph Redmon, Santosh Divvala, Ross Girshick, and Ali Farhadi.
\newblock {You Only Look Once: Unified, Real-Time Object Detection}.
\newblock In \emph{{CVPR}}, 2016.

\bibitem[Ren et~al.(2015)Ren, He, Girshick, and Sun]{ren2015faster}
Shaoqing Ren, Kaiming He, Ross Girshick, and Jian Sun.
\newblock {Faster R-CNN: Towards Real-Time Object Detection with Region Proposal Networks}.
\newblock In \emph{NeurIPS}, 2015.

\bibitem[Ren et~al.(2024)Ren, Huang, Wei, Zhao, Fu, Feng, and Jin]{PixelLM_2024_CVPR}
Zhongwei Ren, Zhicheng Huang, Yunchao Wei, Yao Zhao, Dongmei Fu, Jiashi Feng, and Xiaojie Jin.
\newblock {PixelLM: Pixel Reasoning with Large Multimodal Model}.
\newblock In \emph{{CVPR}}, 2024.

\bibitem[Schuhmann et~al.(2021)Schuhmann, Vencu, Beaumont, Kaczmarczyk, Mullis, Katta, Coombes, Jitsev, and Komatsuzaki]{schuhmann2021laion}
Christoph Schuhmann, Richard Vencu, Romain Beaumont, Robert Kaczmarczyk, Clayton Mullis, Aarush Katta, Theo Coombes, Jenia Jitsev, and Aran Komatsuzaki.
\newblock {LAION-400M: Open Dataset of CLIP-Filtered 400 Million Image-Text Pairs}.
\newblock \emph{arXiv:2111.02114}, 2021.

\bibitem[Schuhmann et~al.(2022)Schuhmann, Beaumont, Vencu, Gordon, Wightman, Cherti, Coombes, Katta, Mullis, Wortsman, et~al.]{schuhmann2022laion}
Christoph Schuhmann, Romain Beaumont, Richard Vencu, Cade Gordon, Ross Wightman, Mehdi Cherti, Theo Coombes, Aarush Katta, Clayton Mullis, Mitchell Wortsman, et~al.
\newblock {LAION-5B: An Open Large-Scale Dataset for Training Next Generation Image-Text Models}.
\newblock \emph{arXiv:2210.08402}, 2022.

\bibitem[Shabtay et~al.(2024)Shabtay, Polo, Doveh, Lin, Mirza, Chosen, Yurochkin, Sun, Arbelle, Karlinsky, et~al.]{vlm_livexiv}
Nimrod Shabtay, Felipe~Maia Polo, Sivan Doveh, Wei Lin, M~Jehanzeb Mirza, Leshem Chosen, Mikhail Yurochkin, Yuekai Sun, Assaf Arbelle, Leonid Karlinsky, et~al.
\newblock {LiveXiv--A Multi-Modal Live Benchmark Based on Arxiv Papers Content}.
\newblock \emph{arXiv:2410.10783}, 2024.

\bibitem[Singh et~al.(2019)Singh, Natarjan, Shah, Jiang, Chen, Parikh, and Rohrbach]{vlm_textvqa}
Amanpreet Singh, Vivek Natarjan, Meet Shah, Yu Jiang, Xinlei Chen, Devi Parikh, and Marcus Rohrbach.
\newblock {Towards VQA Models That Can Read}.
\newblock In \emph{{CVPR}}, 2019.

\bibitem[Sun et~al.(2023)Sun, Fang, Wu, Wang, and Cao]{eva_clip}
Quan Sun, Yuxin Fang, Ledell Wu, Xinlong Wang, and Yue Cao.
\newblock {EVA-CLIP: Improved Training Techniques for CLIP at Scale}.
\newblock \emph{arXiv:2303.15389}, 2023.

\bibitem[Team(2024{\natexlab{a}})]{internlm2}
Intern Team.
\newblock {InternLM2 Technical Report}.
\newblock \emph{arXiv:2403.17297}, 2024{\natexlab{a}}.

\bibitem[Team(2023)]{llama2}
LLaMA Team.
\newblock {LLaMA2: Open Foundation and Fine-Tuned Chat Models}.
\newblock \emph{arXiv:2307.09288}, 2023.

\bibitem[Team(2024{\natexlab{b}})]{qwen2}
Qwen Team.
\newblock {Qwen2 Technical Report}.
\newblock \emph{ArXiv:2407.10671}, 2024{\natexlab{b}}.

\bibitem[Team(2025)]{qwen2_5_vl}
Qwen Team.
\newblock {Qwen2.5-VL Technical Report}.
\newblock \emph{arXiv:2502.13923}, 2025.

\bibitem[Tong et~al.(2024)Tong, Brown, Wu, Woo, IYER, Akula, Yang, Yang, Middepogu, Wang, et~al.]{tong2024cambrian}
Peter Tong, Ellis Brown, Penghao Wu, Sanghyun Woo, Adithya Jairam~Vedagiri IYER, Sai~Charitha Akula, Shusheng Yang, Jihan Yang, Manoj Middepogu, Ziteng Wang, et~al.
\newblock {{Cambrian-1: A Fully Open, Vision-Centric Exploration of Multimodal LLMs}}.
\newblock In \emph{NeurIPS}, 2024.

\bibitem[Tschannen et~al.(2025)Tschannen, Gritsenko, Wang, Naeem, Alabdulmohsin, Parthasarathy, Evans, Beyer, Xia, Mustafa, Hénaff, Harmsen, Steiner, and Zhai]{siglipv2}
Michael Tschannen, Alexey Gritsenko, Xiao Wang, Muhammad~Ferjad Naeem, Ibrahim Alabdulmohsin, Nikhil Parthasarathy, Talfan Evans, Lucas Beyer, Ye Xia, Basil Mustafa, Olivier Hénaff, Jeremiah Harmsen, Andreas Steiner, and Xiaohua Zhai.
\newblock {SigLIP 2: Multilingual Vision-Language Encoders with Improved Semantic Understanding, Localization, and Dense Features}.
\newblock 2025.

\bibitem[Wang et~al.(2021)Wang, Shu, Zhang, Jiang, Wang, Tian, and Wu]{wang2021towards}
Xiao Wang, Xiujun Shu, Zhipeng Zhang, Bo Jiang, Yaowei Wang, Yonghong Tian, and Feng Wu.
\newblock Towards more flexible and accurate object tracking with natural language: Algorithms and benchmark.
\newblock In \emph{CVPR}, 2021.

\bibitem[Wei et~al.(2024)Wei, Tan, Zhong, Yang, and Ma]{wei2024lasagna}
Cong Wei, Haoxian Tan, Yujie Zhong, Yujiu Yang, and Lin Ma.
\newblock {LaSagnA: Language-based Segmentation Assistant for Complex Queries}.
\newblock \emph{arXiv:2404.08506}, 2024.

\bibitem[Wu et~al.(2024)Wu, Zhong, Xing, Lai, Liu, Chen, Wang, Zhu, Lu, Lu, Luo, Qiao, and Dai]{wujiannan_visionLLMv2}
Jiannan Wu, Muyan Zhong, Sen Xing, Zeqiang Lai, Zhaoyang Liu, Zhe Chen, Wenhai Wang, Xizhou Zhu, Lewei Lu, Tong Lu, Ping Luo, Yu Qiao, and Jifeng Dai.
\newblock {VisionLLM v2: An End-to-End Generalist Multimodal Large Language Model for Hundreds of Vision-Language Tasks}.
\newblock In \emph{{NeurIPS}}, 2024.

\bibitem[Wu et~al.(2023)Wu, Zhang, Xu, Jin, Liu, and Loy]{wu2023clim}
Size Wu, Wenwei Zhang, Lumin Xu, Sheng Jin, Wentao Liu, and Chen~Change Loy.
\newblock {CLIM: Contrastive Language-Image Mosaic for Region Representation}.
\newblock \emph{arXiv:2312.11376}, 2023.

\bibitem[Wu et~al.(2019)Wu, Kirillov, Massa, Lo, and Girshick]{wu2019detectron2}
Yuxin Wu, Alexander Kirillov, Francisco Massa, Wan-Yen Lo, and Ross Girshick.
\newblock {Detectron2}, 2019.

\bibitem[XAI.ORG(2024)]{vlm_realworldqa}
XAI.ORG.
\newblock {Grok-1.5 Vision Preview}, 2024.

\bibitem[Xia et~al.(2023)Xia, Han, Han, Pan, Song, and Huang]{Xia2023GSVAGS}
Zhuofan Xia, Dongchen Han, Yizeng Han, Xuran Pan, Shiji Song, and Gao Huang.
\newblock {GSVA: Generalized Segmentation via Multimodal Large Language Models}.
\newblock In \emph{CVPR}, 2023.

\bibitem[Yang et~al.(2023)Yang, Deng, An, Li, Feng, Guo, Yang, and Liu]{yang2023alip}
Kaicheng Yang, Jiankang Deng, Xiang An, Jiawei Li, Ziyong Feng, Jia Guo, Jing Yang, and Tongliang Liu.
\newblock Alip: Adaptive language-image pre-training with synthetic caption.
\newblock In \emph{ICCV}, 2023.

\bibitem[Yang et~al.(2025)Yang, Gu, An, Jiang, Dai, Feng, Cai, and Deng]{yang2025clipcid}
Kaicheng Yang, Tiancheng Gu, Xiang An, Haiqiang Jiang, Xiangzi Dai, Ziyong Feng, Weidong Cai, and Jiankang Deng.
\newblock Clip-cid: Efficient clip distillation via cluster-instance discrimination.
\newblock In \emph{AAAI}, 2025.

\bibitem[Ye et~al.(2022)Ye, Chang, Ma, Shan, and Chen]{OSTrack}
Botao Ye, Hong Chang, Bingpeng Ma, Shiguang Shan, and Xilin Chen.
\newblock Joint feature learning and relation modeling for tracking: A one-stream framework.
\newblock In \emph{ECCV}, 2022.

\bibitem[Yin et~al.(2023)Yin, Fu, Zhao, Li, Sun, Xu, and Chen]{vlm_mme}
Shukang Yin, Chaoyou Fu, Sirui Zhao, Ke Li, Xing Sun, Tong Xu, and Enhong Chen.
\newblock {A Survey on Multimodal Large Language Models}.
\newblock \emph{arXiv:2306.13549}, 2023.

\bibitem[Yu et~al.(2016)Yu, Poirson, Yang, Berg, and Berg]{refcocog}
Licheng Yu, Patrick Poirson, Shan Yang, Alexander~C. Berg, and Tamara~L. Berg.
\newblock {Modeling Context in Referring Expressions}.
\newblock In \emph{{ECCV}}, 2016.

\bibitem[Yuan et~al.(2025)Yuan, Li, Zhang, Huang, Xu, Ji, Tong, Qi, Feng, and Yang]{yuan2025sa2va}
Haobo Yuan, Xiangtai Li, Tao Zhang, Zilong Huang, Shilin Xu, Shunping Ji, Yunhai Tong, Lu Qi, Jiashi Feng, and Ming-Hsuan Yang.
\newblock {Sa2VA: Marrying SAM2 with LLaVA for Dense Grounded Understanding of Images and Videos}.
\newblock \emph{arXiv:2501.04001}, 2025.

\bibitem[Zhai et~al.(2023)Zhai, Mustafa, Kolesnikov, and Beyer]{siglip}
Xiaohua Zhai, Basil Mustafa, Alexander Kolesnikov, and Lucas Beyer.
\newblock {Sigmoid Loss for Language Image Pre-Training}.
\newblock In \emph{{ICCV}}, 2023.

\bibitem[Zhang et~al.(2024{\natexlab{a}})Zhang, Li, Li, Ren, Zou, Liu, Huang, Gao, Leizhang, Li, and Yang]{llava_ground_2024_eccv}
Hao Zhang, Hongyang Li, Feng Li, Tianhe Ren, Xueyan Zou, Shilong Liu, Shijia Huang, Jianfeng Gao, Leizhang, Chunyuan Li, and Jainwei Yang.
\newblock {LLaVA-Grounding: Grounded Visual Chat with Large Multimodal Models}.
\newblock In \emph{{ECCV}}, 2024{\natexlab{a}}.

\bibitem[Zhang et~al.(2024{\natexlab{b}})Zhang, Li, Fei, Yuan, Wu, Ji, Loy, and Yan]{NEURIPS2024_83eb86be}
Tao Zhang, Xiangtai Li, Hao Fei, Haobo Yuan, Shengqiong Wu, Shunping Ji, Chen~Change Loy, and Shuicheng Yan.
\newblock {OMG-LLaVA: Bridging Image-level, Object-level, Pixel-level Reasoning and Understanding}.
\newblock In \emph{{NeurIPS}}, 2024{\natexlab{b}}.

\bibitem[Zhong et~al.(2022)Zhong, Yang, Zhang, Li, Codella, Li, Zhou, Dai, Yuan, Li, et~al.]{regionclip}
Yiwu Zhong, Jianwei Yang, Pengchuan Zhang, Chunyuan Li, Noel Codella, Liunian~Harold Li, Luowei Zhou, Xiyang Dai, Lu Yuan, Yin Li, et~al.
\newblock {RegionCLIP: Region-based Language-Image Pretraining}.
\newblock In \emph{{CVPR}}, 2022.

\bibitem[Zhu et~al.(2024{\natexlab{a}})Zhu, Chen, Shen, Li, and Elhoseiny]{zhu2023minigpt}
Deyao Zhu, Jun Chen, Xiaoqian Shen, Xiang Li, and Mohamed Elhoseiny.
\newblock {Mini{GPT}}-4: Enhancing vision-language understanding with advanced large language models.
\newblock In \emph{{ICLR}}, 2024{\natexlab{a}}.

\bibitem[Zhu et~al.(2024{\natexlab{b}})Zhu, Yanpeng, Wang, Cao, Han, Hou, and Xu]{unit_hangxu}
Yi Zhu, Zhou Yanpeng, Chunwei Wang, Yang Cao, Jianhua Han, Lu Hou, and Hang Xu.
\newblock {UNIT: Unifying Image and Text Recognition in One Vision Encoder}.
\newblock In \emph{{NeurIPS}}, 2024{\natexlab{b}}.

\end{thebibliography}
